\documentclass[10pt,twocolumn,letterpaper]{article}

\usepackage{cvpr}

\usepackage{adjustbox}
\usepackage{array}
\usepackage{booktabs}
\usepackage{colortbl}
\usepackage{wrapfig}
\usepackage{hhline}
\usepackage{multirow}
\usepackage{subcaption} %

\usepackage{amsmath,amsfonts,amssymb}
\usepackage{bm}
\usepackage{nicefrac}
\usepackage{microtype}

\usepackage{changepage}
\usepackage{extramarks}
\usepackage{fancyhdr}
\usepackage{lastpage}
\usepackage{setspace}
\usepackage{soul}
\usepackage{xspace}

\usepackage{url}

\usepackage{enumerate}
\usepackage{enumitem}  %

\usepackage{makecell}

\usepackage{pifont} %

\usepackage{float}
\usepackage{graphicx}
\usepackage{multirow}
 
\usepackage{listings}

\usepackage{algorithm}%
\usepackage{algpseudocode}%

\definecolor{codegreen}{rgb}{0,0.6,0}
\definecolor{codegray}{rgb}{0.5,0.5,0.5}
\definecolor{codepurple}{rgb}{0.58,0,0.82}
\definecolor{backcolour}{rgb}{0.95,0.95,0.92}
\definecolor{red}{rgb}{0.9,0,0}

\lstdefinestyle{mystyle}{
    backgroundcolor=\color{backcolour},   
    commentstyle=\color{codegreen},
    keywordstyle=\color{magenta},
    numberstyle=\tiny\color{codegray},
    stringstyle=\color{codepurple},
    basicstyle=\ttfamily\footnotesize,
    breakatwhitespace=false,         
    breaklines=true,                 
    captionpos=b,                    
    keepspaces=true,                 
    numbers=left,                    
    numbersep=5pt,                  
    showspaces=false,                
    showstringspaces=false,
    showtabs=false,                  
    tabsize=2
}
\usepackage{svg}
\usepackage{enumerate}
\usepackage{amsmath}
\usepackage{txfonts}

\newcolumntype{L}[1]{>{\raggedright\let\newline\\\arraybackslash\hspace{0pt}}m{#1}}
\newcolumntype{C}[1]{>{\centering\let\newline\\\arraybackslash\hspace{0pt}}m{#1}}
\newcolumntype{R}[1]{>{\raggedleft\let\newline\\\arraybackslash\hspace{0pt}}m{#1}}

\newcommand{\ignore}[1]{}

\DeclareMathAlphabet{\mathbfit}{OML}{cmm}{b}{it}

\makeatletter
\DeclareRobustCommand\onedot{\futurelet\@let@token\@onedot}
\def\@onedot{\ifx\@let@token.\else.\null\fi\xspace}

\makeatother

\definecolor{MyDarkBlue}{rgb}{0,0.08,1}
\definecolor{MyAqua}{rgb}{0,0.7,0.7}
\definecolor{MyDarkGreen}{rgb}{0.02,0.6,0.02}
\definecolor{MyDarkRed}{rgb}{0.8,0.02,0.02}
\definecolor{MyDarkOrange}{rgb}{0.40,0.2,0.02}
\definecolor{MyPurple}{RGB}{111,0,255}
\definecolor{MyRed}{rgb}{1.0,0.0,0.0}
\definecolor{MyGold}{rgb}{0.75,0.6,0.12}
\definecolor{MyDarkgray}{rgb}{0.66, 0.66, 0.66}

\usepackage{caption}
\DeclareCaptionLabelFormat{cont}{#1~#2\alph{ContinuedFloat}}
\usepackage{sidecap}
\usepackage{pifont}

 \newcommand{\galleryRowCompare}[5]{
   \vspace*{-0.075cm}
   \includegraphics[width=0.2\textwidth]{#1/#2} &
   \includegraphics[width=0.2\textwidth]{#1/#3} &
   \includegraphics[width=0.2\textwidth]{#1/#4} &
   \includegraphics[width=0.2\textwidth]{#1/#5} &

   \tabularnewline 
 }

\newcommand{\projectname}[0]{Infinigen Indoors}

\usepackage{graphicx}
\usepackage{amsmath}
\usepackage{amssymb}
\usepackage{booktabs}

\newcommand{\parbf}[1]{\par \noindent \textbf{#1}.}

\definecolor{cvprblue}{rgb}{0.21,0.49,0.74}
\usepackage[pagebackref,breaklinks,colorlinks,citecolor=cvprblue]{hyperref}
\usepackage{graphicx}
\usepackage[accsupp]{axessibility} 
\usepackage{enumerate}

\DeclareMathOperator*{\argmin}{arg\,min}

\title{Infinigen Indoors: Photorealistic Indoor Scenes using Procedural Generation}

\author{
Alexander Raistrick$^*$,\, Lingjie Mei$^*$,\, Karhan Kayan$^*$, {\small($^*$equal contribution; random order)} \\
David Yan,\, 
Yiming Zuo,\,
Beining Han,\, 
Hongyu Wen,\,
Meenal Parakh,\\
Stamatis Alexandropoulos,\,
Lahav Lipson,\,
Zeyu Ma,\,
Jia Deng \\
Department of Computer Science, Princeton University\\
}

\begin{document}
\maketitle

\begin{abstract}

We introduce Infinigen Indoors, a Blender-based procedural generator of photorealistic indoor scenes. It builds upon the existing Infinigen system, which focuses on natural scenes, but expands its coverage to indoor scenes by introducing a diverse library of procedural indoor assets, including furniture, architecture elements, appliances, and other day-to-day objects. It also introduces a constraint-based arrangement system, which consists of a domain-specific language for expressing diverse constraints on scene composition, and a solver that generates scene compositions that maximally satisfy the constraints. We provide an export tool that allows the generated 3D objects and scenes to be directly used for training embodied agents in real-time simulators such as Omniverse and Unreal. Infinigen Indoors is open-sourced under the BSD license. Please visit \href{https://infinigen.org}{infinigen.org} for code and videos.

\end{abstract}

\section{Introduction}
\label{sec:intro}

Synthetic data rendered by conventional computer graphics has seen increasing adoption in computer vision\cite{gta5koltun2016playing, crestereo, law2022label, li2022practical, lipson2022coupled, ma2022multiview, deitke2022} and AI research\cite{thambawita2022singan, guibas2017synthetic,hurl2019precise}, especially for 3D vision\cite{teed2021droid, teed2022deep, lipson2021raft, teed2020raft, wang2021tartanvo, haugaard2022surfemb, teed2021raft} and embodied AI\cite{ai2thor, szot2021habitat, srivastava2022behavior,xiang2020sapien,james2020rlbench,dosovitskiy2017carla}. Synthetic data can be rendered in unlimited quantities and can automatically provide high-quality 3D ground truth, enabling large-scale training of computer vision models and embodied agents. Notably, many state-of-the-art 3D vision systems\cite{Teed2021DROIDSLAMDV, teed2022deep} and robotic systems \cite{Lee2020LearningQL,KaufmannChampion} have been trained purely in simulation yet perform surprisingly well in the real world \emph{zero-shot}. 

A promising direction for creating synthetic data is procedural generation, which uses mathematical rules to create 3D objects and scenes, as opposed to manual sculpting or real-world scanning. These mathematical rules can have parameters that are randomized to allow infinite variations. For example, trees can be generated through a recursive set of rules that randomly branch off. Compared to reusing a fixed, static set of 3D assets, procedural generation can greatly improve the diversity of the synthetic data and the simulated environments. 

Infinigen~\cite{infinigen2023infinite} is a recent work that pushed the idea of procedural generation to the limit. Infinigen is an open-source system that generates photorealistic 3D scenes \emph{fully procedurally}, meaning that every 3D asset, from shape to material, from large structures to small details, is completely procedural, without using any external static asset. Being fully procedural means that every aspect of the 3D scene, from the details of individual objects to their arrangements in a scene, can be customized and controlled by simply modifying the underlying mathematical rules. As a result, a 3D scene can be randomized at all levels down to the smallest details, as opposed to only at the level of object arrangement, which was common in earlier work that used procedural generation. 

However, the current Infinigen system is limited to natural scenes and objects (terrains, animals, plants, etc.\@). Although natural scenes could be sufficient for training foundation models as evidenced by natural evolution \cite{infinigen2023infinite}, this hypothesis remains unproven and may require additional advances in learning algorithms and architecture designs. Evidence from the current literature suggests that synthetic training data that more closely approximates the application domain is still likely to lead to better downstream performance. 

\begin{figure*}[!h]
\includegraphics[width=\linewidth]{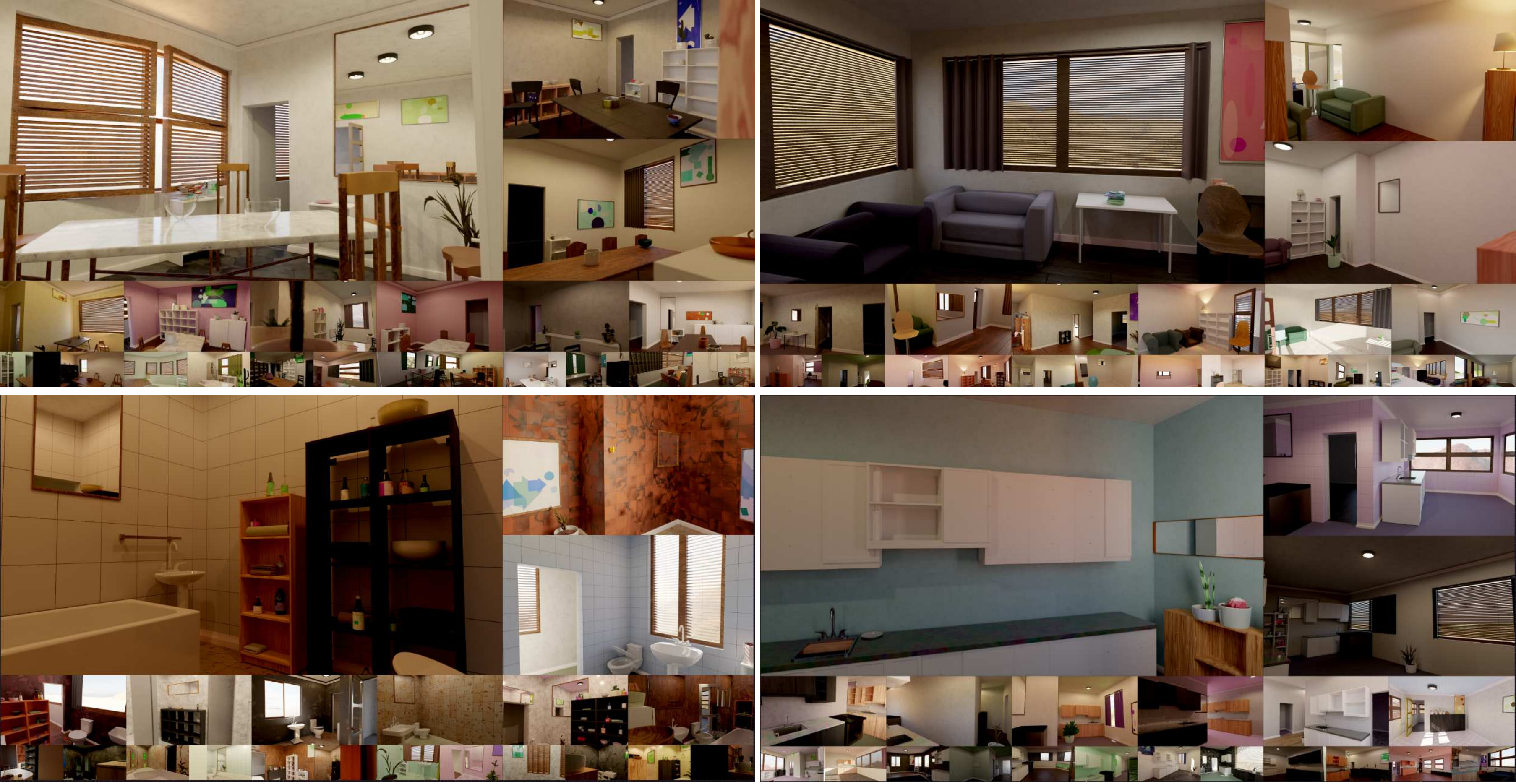}
 \caption{Random, \textit{non cherry-picked} sample of images generated by our system. From top left to bottom right, we show images from dining rooms, bathrooms, living rooms and kitchens. Please see Appendix \ref{sec:rand_sample} for an extended random sample. }
\label{fig:teaser}
\vspace{-1.5em}
\end{figure*}

\begin{figure}[]
    \centering
  \resizebox{\linewidth}{!}{
    \includegraphics[width=\textwidth]{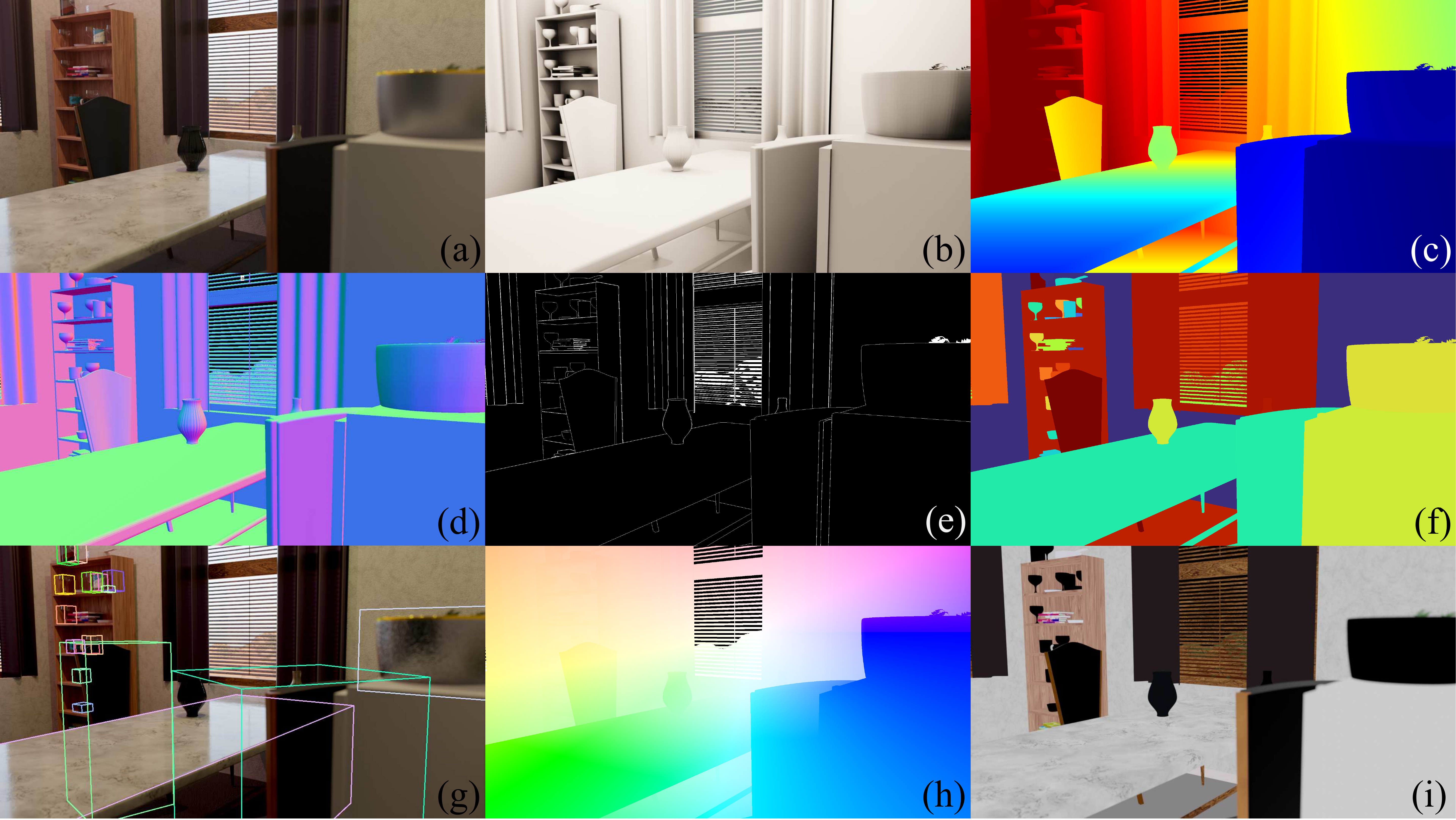}}
    \caption{Each image (a) is rendered from a mesh (b), from which we can also extract Depth (c), Surface Normals (d), Occlusion Boundaries (e), Segmentation (f), Bounding Boxes (e) and Optical Flow (h), with Albedo (i) from rendering metadata.}
    \label{fig:ground_truth}
    \vspace{-1.5em}
\end{figure}

To overcome this limitation, we introduce Infinigen Indoors, a procedural generator of photorealistic indoor scenes. It expands the coverage of Infinigen to indoor scenes, which are relevant for many high-impact applications including robotics and augmented reality. Infinigen Indoors generates diverse indoor objects, including furniture, appliances, cookware, dining utensils, architectural elements, and other common day-to-day objects. It also generates full indoor scenes, including the interior of multi-room, multi-floor buildings, with object arrangements that are physically and semantically plausible. Fig.~\ref{fig:teaser} shows random samples of generated scenes, and Fig.~\ref{fig:ground_truth} shows some automatic annotations. 

Like the original Infinigen, Infinigen Indoors is not a fixed set of 3D models or scenes; instead, it is an open-source generator that can create unlimited variations both at the object level and at the scene level. Infinigen Indoors is also 100\% procedural, using no external assets and using only mathematical rules to generate everything from scratch. 

Infinigen Indoors builds upon the original Infinigen and Blender \cite{blender} but makes significant new contributions. The main contributions include (1) a library of procedural generators of indoor assets, (2) a constraint-based arrangement system, (3) a tool to export the generated scenes to real-time simulators such as NVIDIA Omniverse ~\cite{omniverse} and Unreal Engine\cite{unrealengine}.

\begin{table*}[t]
    \centering
    \resizebox{\linewidth}{!}{%
    \begin{tabular}{l|c|c|c|c|c|c|l} 
        \toprule
            \multirow{2}{*}{ Dataset} & Arrangement & Procedural & Provides & \# Scenes & \# Assets & Free & \multirow{2}{*}{External Asset Source}  \\ %
        & Method & Assets &  Procedural Code & in Total & in Total &  Assets  &\\
        \midrule
        3DSSG~\cite{3DSSG2020} & Real-world scans & No & N/A & 1.5K & 48K & Yes & 3RScan \cite{Wald2019RIO}  \\ 
        Matterport3D~\cite{chang2017matterport3d} &  Real-world scans & No & N/A & 2K \cite{patil_advances_2023} & - & Yes & None\\ 
        Stanford 2D-3D-S~\cite{2017arXiv170201105A} & Real-world scans & No & N/A & 270 & - & Yes & None \\ 
        ScanNet~\cite{dai2017scannet} & Real-world scans & No & N/A & 1.5K & - & Yes & None \\ 
        SceneNN~\cite{hua2016scenenn} & Real-world scans & No & N/A & 100 & - & Yes & None \\ 
        OpenRooms~\cite{Li2020OpenRoomsAO} & Real-world scans & No & No & 1.3K & 3K \cite{infinigen2023infinite} & No (\$500) & ShapeNet~\cite{shapenet}, Scan2CAD~\cite{avetisyan2019scan2cad}, Adobe Stock~\cite{adobestock} \\ 
        Replica~\cite{replica19arxiv} &  Real-world scans & No & N/A & 18 & - & Yes & None \\ 
        Structured3D~\cite{avetisyan2019scan2cad} & Artist layouts & No & N/A & 22K & 472K & No & Professional Designers \\ 
        Hypersim~\cite{Roberts_2021_ICCV} & Artist layouts & No & N/A & 461 & 59K & No (\$6000) & Evermotion Architectures~\cite{evermotionarch}  \\ 
        InteriorNet~\cite{li2018interiornet} & Artist layouts & No & N/A & 22M & 1M & No  & Manufactures / Kujiale~\cite{kujiale} \\ 
        Habitat 3.0~\cite{khanna2023habitat} & Artist layouts & No & N/A & 211 & 18.7K & Yes & Floorplanner~\cite{floorplanner}, Proffesional Designers \\ 
        3D-FRONT~\cite{fu20213d} &  Artist layouts & No & N/A & 19K & 13K & Yes & 3D-FUTURE~\cite{3dfuture}  \\ 
        Robotrix~\cite{garcia2018robotrix} &  Artist layouts & No & N/A & 16 & - & No & UE4Arch~\cite{UE4Arch}, UnrealEngine Marketplace~\cite{unrealenginemarket}\\ 
        DeepFurniture~\cite{liu2019furnishing} & Artist layouts & No & N/A & 20K & - & No & Adobe Mixamo~\cite{adobemixamo} \\         
        SceneNetRGBD~\cite{zheng2020structured3d} & Obj. Cat. Dist. & No & N/A & $\infty$ & 5.1K & Yes & SceneNet~\cite{handa2016scenenet}, ShapeNet~\cite{shapenet} \\ 
        LUMINOUS~\cite{zhao2021luminous} & Hierarchical Sampling & No & Yes & $\infty$& 2K& Yes & AI2-THOR~\cite{ai2thor} \\ 
        SceneNet~\cite{7487797} & Optimizer & No & No & $\infty$& 3.7K & Yes &  3DModelFree~\cite{3dmodelfree}, ModelNet~\cite{wu20153d}, Archive3D~\cite{archive3d}, Stanford database \\ 
        ProcTHOR~\cite{deitke2022} & Procedural Rules & No & Yes & $\infty$& 1.6K & Yes & AI2-THOR~\cite{ai2thor}, Professional Designers \\ 
        Holodeck~\cite{yang2024holodeck} &  LLM & No & Yes & $\infty$ & 50K & Yes & Objaverse~\cite{deitke2022objaverse} \\ 
        Aria~\cite{aria} & Procedural Rules & No & No & 100K & 8K & - & - \\ 
        \midrule
        \projectname{} (Ours) & Constraint Language & Yes & Yes & $\infty$ & $\infty$& Yes & None \\
        \bottomrule
    \end{tabular}}

\caption{Comparisons to existing datasets and generators. Many existing datasets/generators use external, static asset libraries and have limited number of scenes. Ours is fully procedural, without using any external source.  Dashes represent numbers we could not acquire or estimate.}

\label{tab:dataset_comparisons}
\vspace{-1.5em}
\end{table*}

Our second contribution---a constraint-based arrangement system---offers a new capability specifically targeting indoor settings. Indoor scenes are artificial, and object arrangement exhibits a greater degree of regularity than natural scenes: for example, furniture usually does not block the entrance of a room. We thus develop a system that lets the user specify scene arrangement constraints through a domain-specific language using a set of Python APIs. The constraints cover many types of common arrangement, including symmetry (``{\it Place chairs symmetrically around the table}''), spatial relation (``{\it Place plant pots close to windows} ''), quantity (``{\it An equal number of knives and forks}''), physics (``{\it Ensure vases do not overhang}''), and accessibility (``{\it Ensure there is free-space in-front of all appliances}''). The constraints can be understood as a type of declarative procedural rules that express what the user desires but not how to achieve it. Like other procedural rules, the constraints can be randomized and can be customized by the user.

In addition to constraint specification, our arrangement system also includes a \emph{constraint solver}, which searches for an arrangement that maximally satisfies a set of given constraints. Our solver greedily performs simulated annealing on whole-house floor plans, followed by large furniture layouts and then small objects. Compared to existing approaches for scene arrangement, our solver is highly expressive, supporting complex compositional constraints that are challenging or infeasible for existing approaches. In addition, it is the first solver integrated with an open-source and fully procedural generator. 

Our constraint-based arrangement system is a significant contribution because it vastly improves the generation system's usability and customizability. Because it separates constraint specification from constraint solving, a user can conveniently express the objectives of procedural generation without worrying about implementation. This capability is not available in the original Infinigen, where the user has to customize the procedure rules at the implementation level. 

Our third contribution---exporting to real-time simulators---is also noteworthy because it allows the generated 3D objects and scenes to be directly used for training embodied agents in real-time simulators such as Omniverse. Thus, Infinigen Indoors can supply diverse 3D assets for simulation environments and enhance their domain randomization. 

To validate the effectiveness of the generated data and demonstrate our system's unique customizability, we use Infinigen Indoors to generate synthetic data for shadow removal and occlusion boundary detection, two tasks that lack abundant existing training data. Our experiments show that data from our system improves generalization performance on indoor scenes.  

Like the original Infinigen, Infinigen Indoors will be open-sourced under the BSD license to enable free and unlimited use by everyone, and to enable community contributions of additional procedural generators.

\section{Related Work}

We provide a detailed comparison of Infinigen Indoors with existing datasets and generators in Tab.~\ref{tab:dataset_comparisons}.  

\parbf{Real-world datasets} Various real-world datasets have been introduced for indoor scene understanding~\cite{3DSSG2020,6130298,Song_2015_CVPR,2017arXiv170201105A,chang2017matterport3d,hua2016scenenn,dai2017scannet}, including the earlier and widely used NYUv2~\cite{6130298} and Sun RGB-D~\cite{Song_2015_CVPR}, as well as more recent datasets~\cite{3DSSG2020, 2017arXiv170201105A, dai2017scannet, arkitscenes}. However, real-world datasets are labor-intensive to collect and limited in size. In addition, real-world 3D ground truth can be difficult to acquire due to the limitations of depth sensors, which include limited resolution and range, errors with transparent/reflective surfaces, and artifacts at object edges. 

\parbf{Synthetic Indoor Datasets} There are many existing synthetic datasets for indoor scenes~\cite{replica19arxiv, avetisyan2019scan2cad, li2018interiornet, khanna2023habitat, fu20213d, ai2thor, garcia2018robotrix, liu2019furnishing, Roberts_2021_ICCV, 7487797}. However, the underlying 3D assets of many datasets are not  freely accessible, limiting their utility. In addition, most use a \emph{static} library of 3D assets, limiting their diversity. Recent work ~\cite{zhao2021luminous, deitke2022} has incorporated procedural generation for scene layout and floor plan generation, but still relies on static libraries of objects and materials. In contrast, \projectname{} is 100\% procedural, with all assets from shape to texture generated from scratch with unlimited variation. 

\parbf{Object arrangement and layout generation}

Constraints are potent tools to describe the layout of a scene. Early works like \cite{Xu2002ConstraintBasedAP} represent constraints as hard-coded programs, and \cite{McCormac2017SceneNetRC} represent them as physical relations. Data-driven works like \cite{atiss, scene_former, ritchie_fast_2018, planit, layoutenhancer, para_generative_2020} learn constraints implicitly from data. Such implicit constraints are less customizable, interpretable, and controllable than Infinigen Indoors.
Recently, modeling constraints using probabilistic graphs have become more popular: \cite{wallgrid} uses pairwise grouping, while \cite{makeithome, Deitke2022ProcTHORLE} further extends it to spatial and hierarchical constraints. \cite{revJumpMCMC} uses factor graphs to parse the constraints, while \cite{Merrell2010ComputergeneratedRB} models them with Bayesian network. \cite{QiHumanCentric, Zhao2021LUMINOUSIS} formulates constraints as potentials Markov Random Fields on a fixed graph, which capture only non-compositional and associative constraints for rooms and objects.

Compared to existing systems, ours is the first to integrate directly with procedural object generators, and our constraint language is higher level and more easily extendable than existing systems. Our system specifies high level goals for abstract classes of objects (e.g. 'furniture', 'storage'), rather than exhaustive distance/angle distributions for specific objects \cite{makeithome} which must be fitted to example scenes. Our language also supports compositional constraints such as ``place glassware only on shelves against a dining-room wall.'' These features allow users to write new constraints for their specific needs, including domains without existing artist-made scenes. Our constraint solver uses simulated annealing, following prior work \cite{makeithome}, but involves moves that are unique to our constraint language, including updates to object-object relations or changing the parameters of procedural objects (e.g. the size of a table). 

\section{Method}

\subsection{Procedural Asset Generation}

\begin{figure}[]
    \centering
  \resizebox{\linewidth}{!}{
    \includegraphics[width=\textwidth]{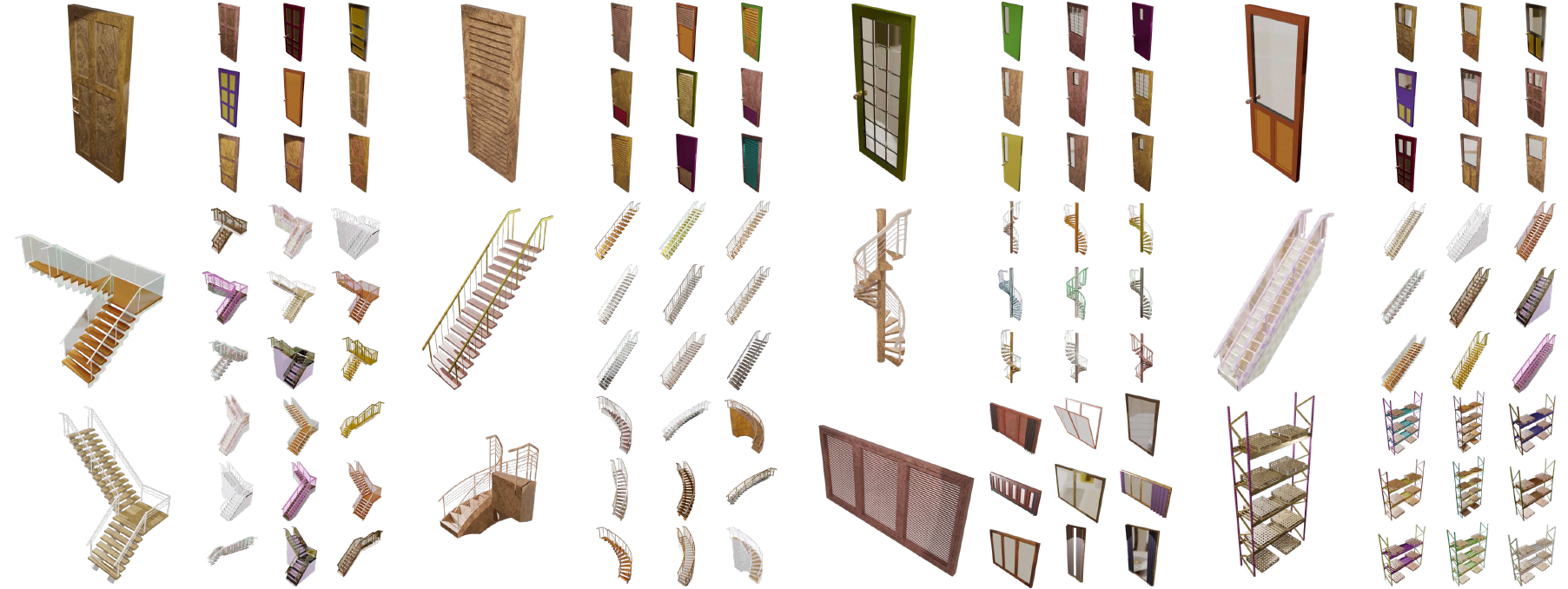}}
    \caption{Random samples of procedurally generated doors (top), staircases (middle/bottom) and windows/warehouse shelving (bottom-right).}
    \label{fig:elements}
    \vspace{-1.5em}
\end{figure}

\begin{figure}
    \centering
  \resizebox{\linewidth}{!}{
    \includegraphics[width=\textwidth]{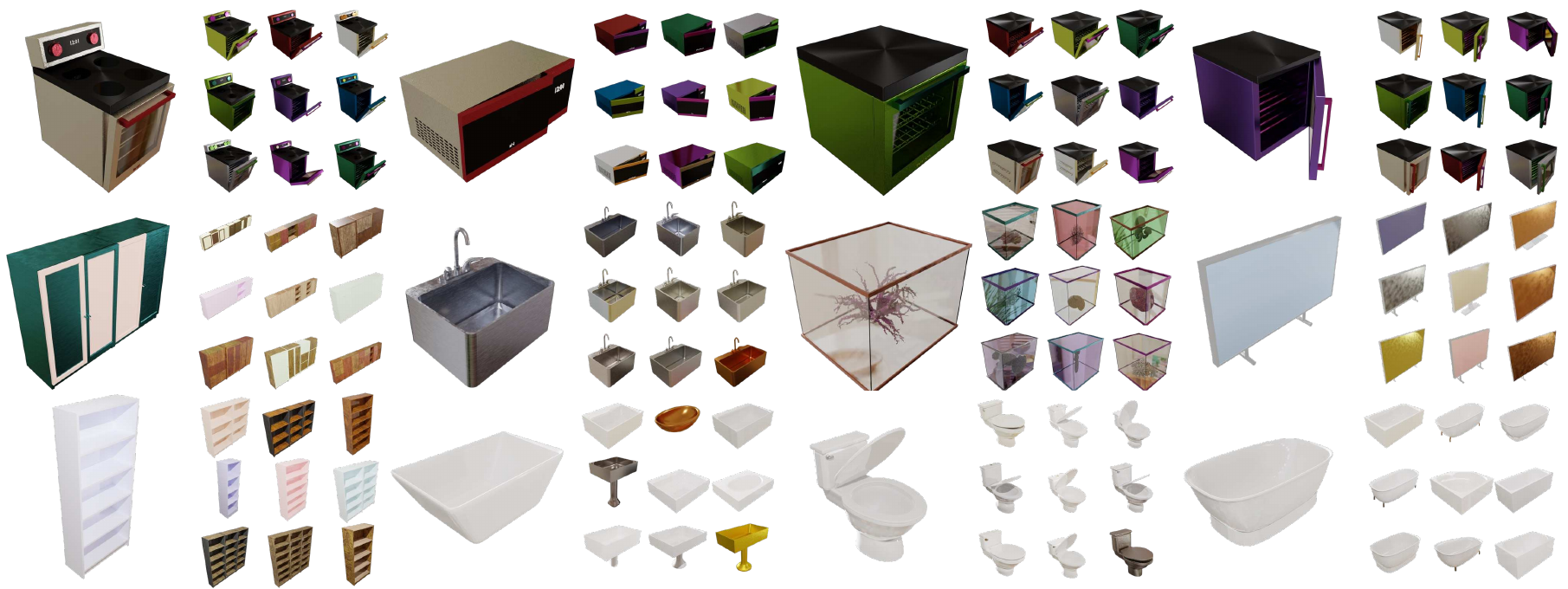}}
    \caption{Random samples of procedurally generated ovens, dishwasher and sinks (top/middle), living-room furniture (middle) and bathroom fixtures (bottom).}
    \label{fig:appliances}
    \vspace{-1.5em}
\end{figure}

\begin{figure}
    \centering
  \resizebox{\linewidth}{!}{
    \includegraphics[width=\textwidth]{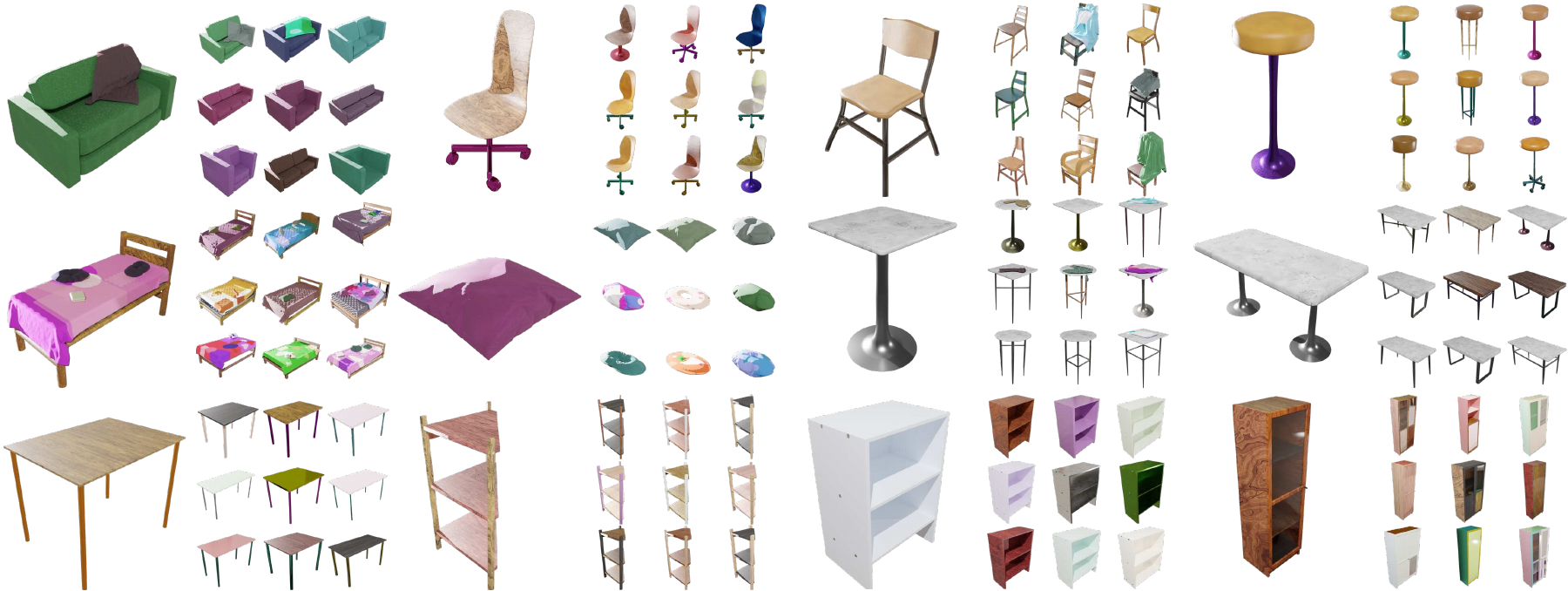}}
    \caption{Random samples of procedurally generated furniture, including sofa, chairs, and beds (top), tables (middle/bottom), and shelves (bottom).}
    \label{fig:seatings}
\end{figure}

\begin{figure}
    \centering
  \resizebox{\linewidth}{!}{
    \includegraphics[width=\textwidth]{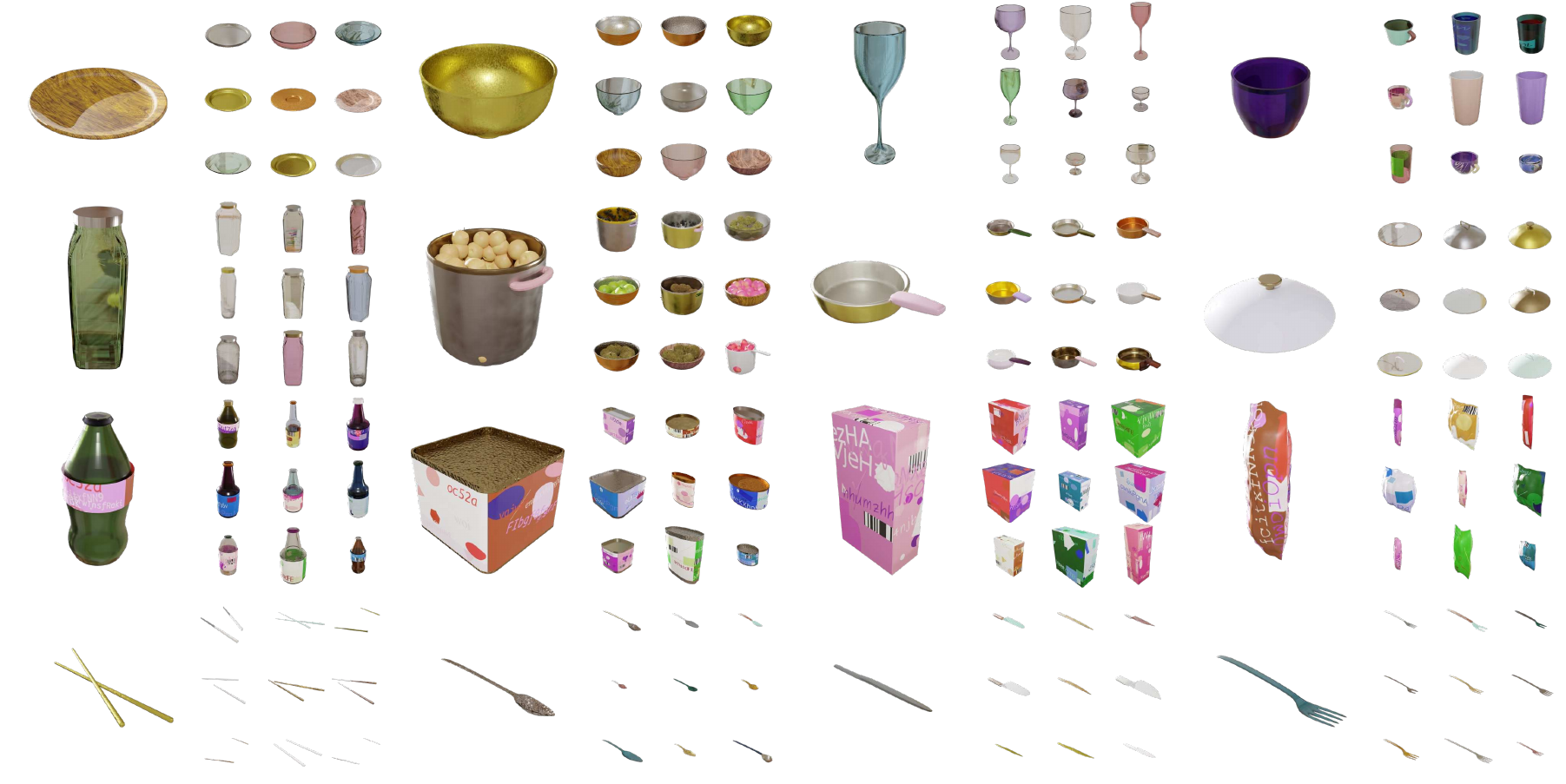}}
    \caption{Random samples of procedurally generated tableware, including 
    dinnerware (Row 1-2), cookware (Row 2), food containers (Row 3) and dining utensils (Row 4).}
    \label{fig:tableware}
\end{figure}

\begin{figure}
    \centering
  \resizebox{\linewidth}{!}{
    \includegraphics[width=\textwidth]{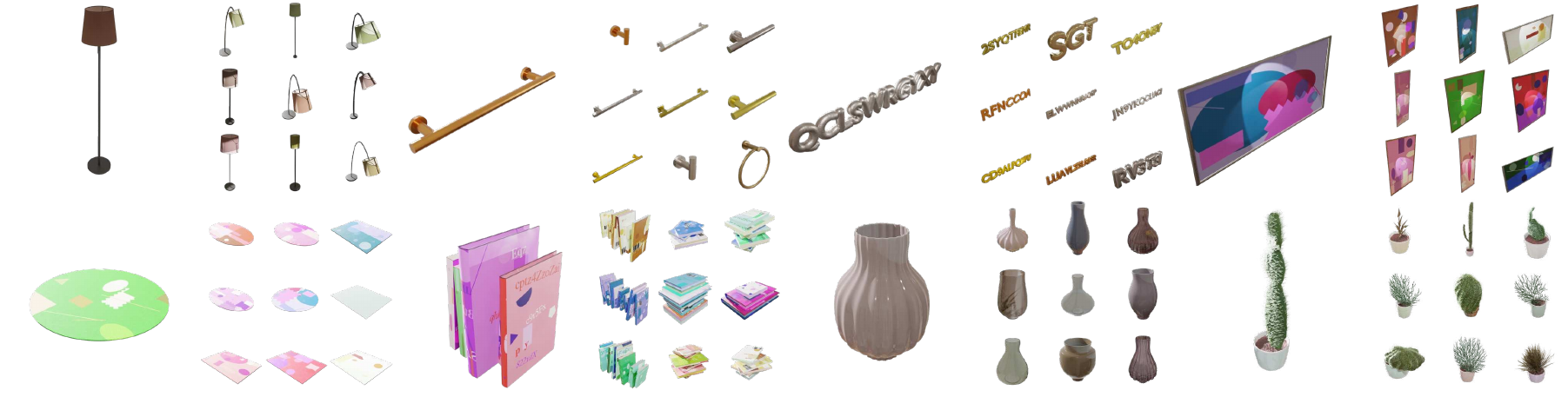}}
    \caption{Random samples of procedurally generated home decorations, including lamps, hardware, balloons, wall decor (top), rugs, book stacks, vases, and plants (bottom). 
    Small assets for decoration purposes, usually attached to the ground or walls.}
    \label{fig:misc}
    \vspace{-1.5em}
\end{figure}

\begin{figure} [t]
    \centering
  \resizebox{\linewidth}{!}{
    \includegraphics[width=\textwidth,trim={0 .5cm 0 0},clip]{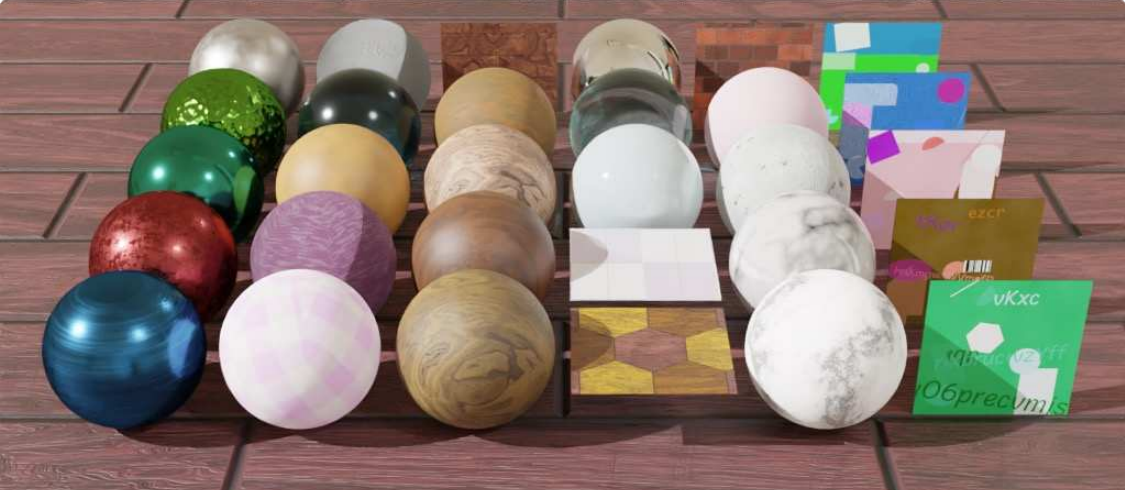}}
    \resizebox{\linewidth}{!}{
    \includegraphics[width=\textwidth]{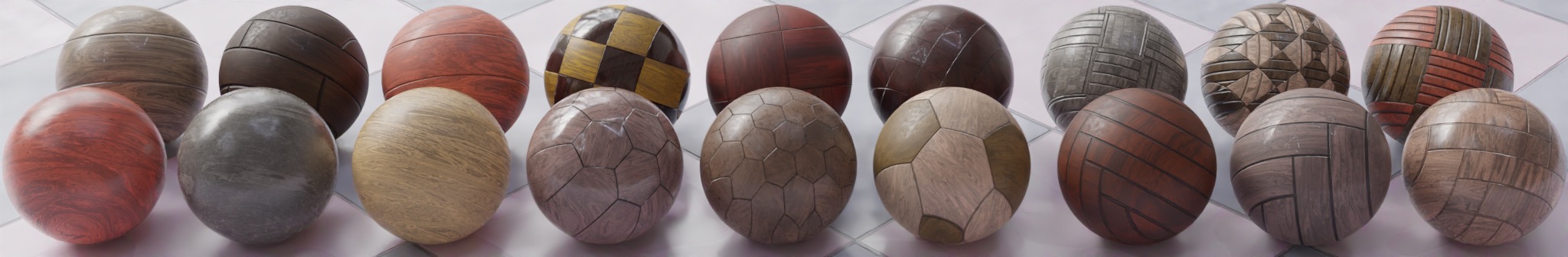}}
    \caption{A collection of materials generated in \projectname{}. The first figure shows one material per generator, with columns (1-3) used on assets of various sizes, (4-5) used on assets and rooms,  and  (6) for abstract art and text. The second figure shows multiple materials from the same generator with different parameters.}
    \label{fig:materials}
\end{figure}

\begin{figure} [t]
    \centering
  \resizebox{\linewidth}{!}{
    \includegraphics[width=\linewidth]{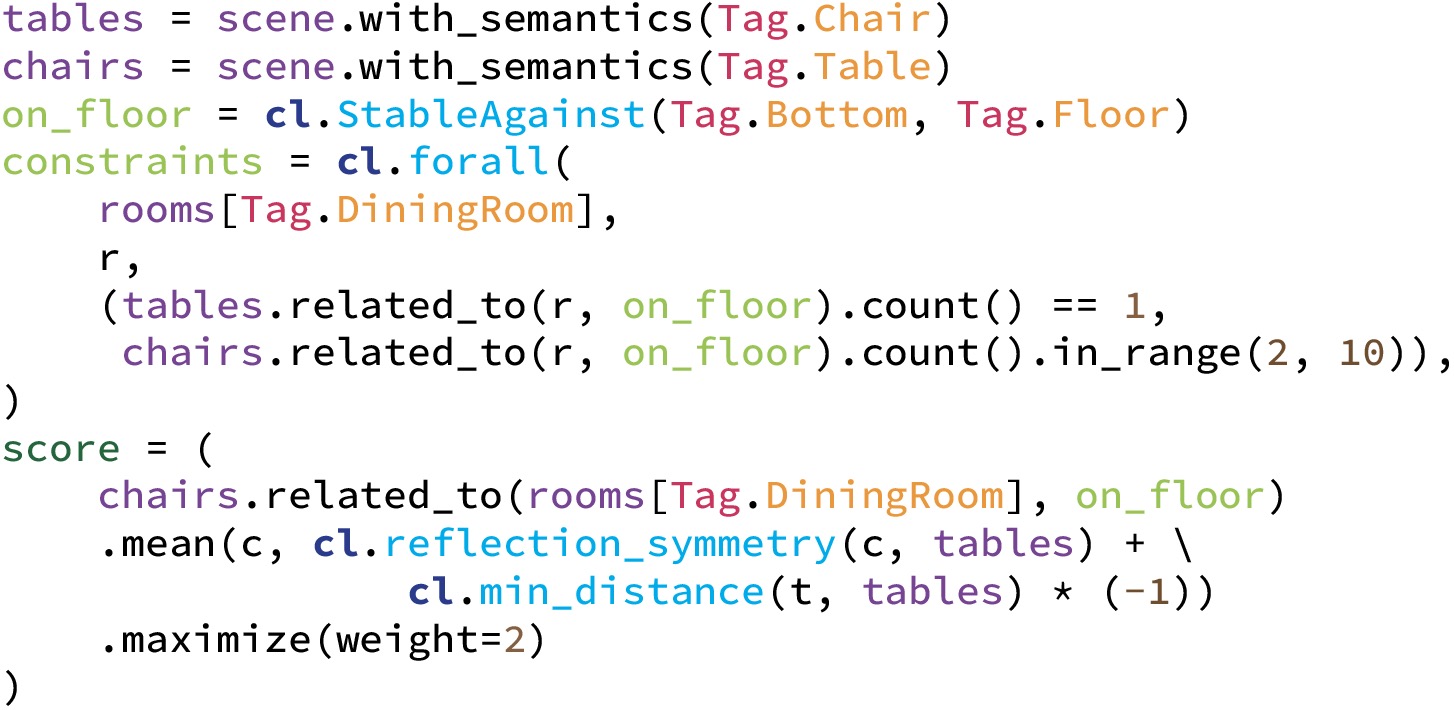}}
    \caption{Example usage of our constraint specification API, specifying the quantity and aesthetic constraints for a dining table and chairs.}
    \label{fig:constraint_excerpt}
\end{figure}

\begin{figure*}[th!]
    \centering
    \includegraphics[width=\textwidth, trim={1mm 1mm 1mm 1mm}, clip]{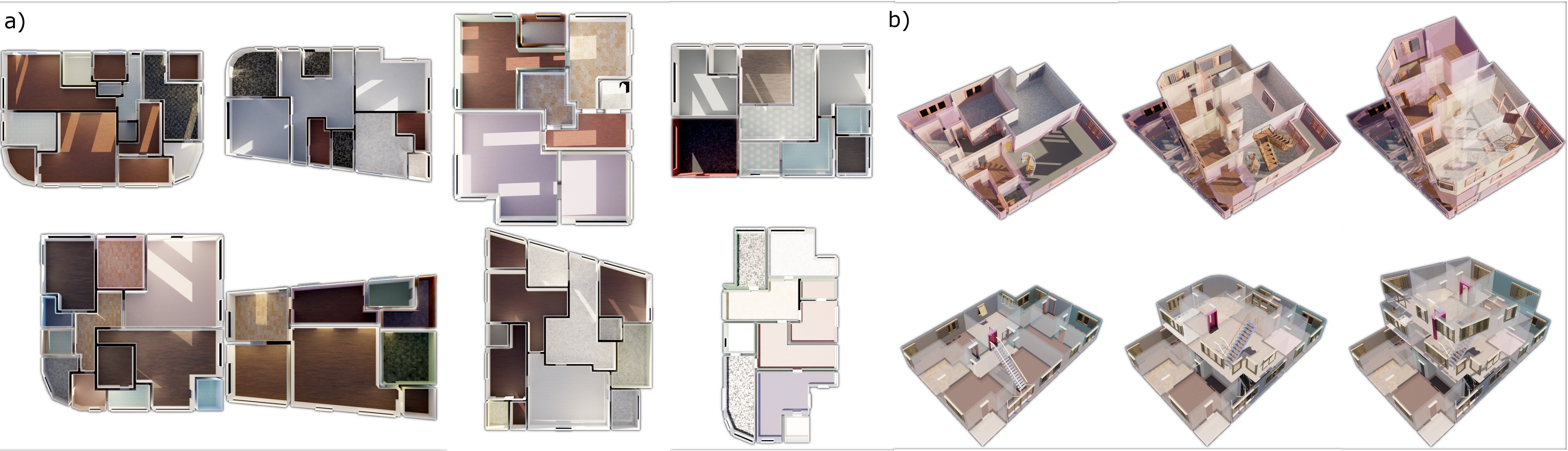}
    \caption{In Fig. a), we show ten randomly generated single-story floor plans with a diverse set of room combinations, connectedness, and overall contours. In Fig. b), we show results for the generation of multistory floor plans. Floors 0,1,2 are displayed separately. Staircases connect adjacent floors. We remove exterior walls and ceilings for visibility.}
    \label{fig:floor_plan}
\end{figure*}

\begin{figure*}[th!]
\resizebox{\linewidth}{!}{\includegraphics[width=\linewidth]{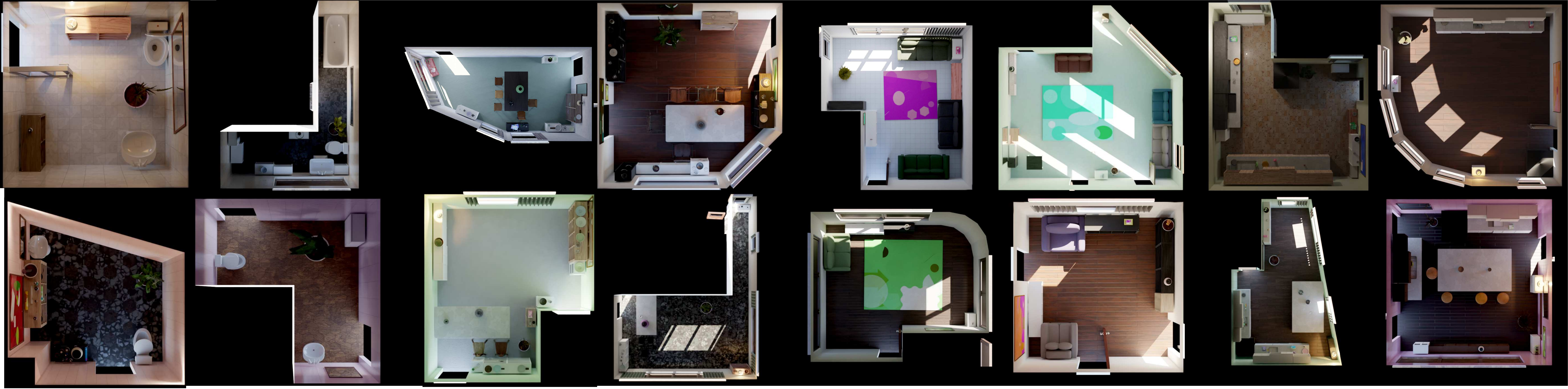}}
 \caption{Qualitative room arrangement results, grouped by room type. From left to right, we show bathrooms, dining rooms, living rooms and kitchens.} 
\label{fig:room_final}
\vspace{-1.5em}
\end{figure*}

All assets used in \projectname{} are generated from scratch by compact probabilistic programs. These programs have many human-controllable parameters, which are randomized by default, or can be manually overridden by the user. These parameters are used along with additional low-level random noise to generate meshes via geometry nodes, modifiers, or mesh manipulations in Blender. We provide a total of 79 randomized procedural object generators. By category, we cover Appliances (10 generators, 112 params), Windows/Doors/Staircases (14 generators, 127 params), Furniture (17 generators, 216 params), Decorations (15 generators, 92 params), and Small Objects (19 generators, 194 params). See Appendix \ref{sec:asset_gen} for a list. 

\noindent \textbf{Architectural elements} shown in Fig.~\ref{fig:elements} are integrated into room as fixtures. We use array repetition of atomic components to build staircases, and mesh booleaning to cut out the panels of doors and windows. 

\noindent \textbf{Large objects} shown in Fig.~\ref{fig:appliances} and \ref{fig:seatings} provide assets related to cooking, seating and storage. We use soft-body collision simulation to model soft blankets, clothing, and stuffed pillows when they are put on a supporting surface.

\noindent \textbf{Small objects} shown in Fig.~\ref{fig:tableware} and \ref{fig:misc} can be attached to support surfaces, walls, or ceilings.  We also devised a combined text-and-shape logo generator that produces procedural texture for fabrics, food packaging, and art decor. We use cloth simulation to inflate balloons and food packaging with air. 

\noindent \textbf{Materials} are all procedurally created with Blender's shader nodes, as shown in Fig.~\ref{fig:materials}. We provide 30 material generators with 120 controllable parameters in total, split approximately evenly between types of wood, ceramic, fabric, metal, and others. We cover 78\% of OpenSurfaces'\cite{bell13opensurfaces} material categories, up from 21\% for Infinigen.
\subsection{Constraint Specification API}

Indoor scene layouts are highly regular, and follow complex rules governing ergonomics, aesthetics, and functionality. Moreover, the rules that apply to a particular object depend on context - for example, tables are placed against walls when used as desks in study rooms, but must be far from walls and surrounded by chairs when used in a dining room. To capture this, we provide a high-level Constraint Specification API, which allows the user to write expressive objective functions to describe the properties of a desirable scene. An example of the Constraint Specification API is shown in \ref{fig:constraint_excerpt}.

Each constraint in our API is a compute graph of geometric, set filtering, and arithmetic operations. Geometric operators are designed to compute spatial and geometric properties, including {\it minimum distance}, {\it rotational and reflection symmetry}, {\it angle alignment}, {\it 2D free-space}, {\it accessibility} and {\it volume} or {\it area} of objects. Each geometric operator accepts a set of objects as input, which can be provided by filtering the scene using semantics and scene graph relations. This allows the user to create scoped constraints that apply only to objects attached to specific surfaces or rooms. Additionally, these geometry terms are affected by the parameters (length, width, etc.) of the procedurally generated assets, meaning that optimizers can automatically discover optimal furniture parameters given available space and constraints. Our system features common scalar arithmetic operations, comparisons, and \texttt{forall} / \texttt{sum} operators to gather results over sets of objects. Please see Appendix \ref{sec:api_desc} for the full API and more examples of constraint specifications. 

A more concrete example of our constraint language can be found in Fig.~\ref{fig:constraint_excerpt}, where the constraint program specifies common-sense human ergonomics and semantic relations found in residential homes. This constraint graph has a total of 1058 nodes, which compute 11 hard constraints and 25 score terms (soft constraints). We provide example constraint specifications for living rooms, bathrooms, dining rooms, kitchens, and warehouses. 

We believe that many users will consider creating custom constraints tailored to particular applications when generating training data. Our constraint system is designed to allow easy customization. Our initial spec.\@ has avg.\@ 15 constraints specific to each room: approx.\ 15 lines of Python. We believe this cost is very tractable when users need customization. 

\subsection{Arrangement Solver}
\label{sec:solver}
Because our Constraint Specification API is flexible, the solver needs to search a prohibitively large space in which finding an exact minimum is impossible. To deal with this, it uses Simulated Annealing\cite{simulated_annealing} with Metropolis-Hastings criterion \cite{Metropolis1953EquationOS, hastings_monte_1970}. The solver first takes the current state $s$ and randomly chooses a move category. 
It then uses the constraint graph to generate a proposed state $s'$ that can be reached using the move. The current and proposed states are evaluated on the graph specified by the provided constraints and score terms, yielding loss terms $l(s)$ and $l(s')$. Then, the solver calculates the transition probability between $s$ and $s'$ as
\[ p(s'| s) = \min \left [\exp \left (\frac{l(s) - l(s')}{\tau}\right ), 1 \right ]\]
where $\tau$ is the temperature of the solver, which cools exponentially from $\tau=0.25$ to $\tau=0.001$.

Our solver allows both discrete and continuous moves:

\noindent \textbf{Addition} - Adds a procedural object to the scene. 

\noindent \textbf{Deletion} - Deletes an object from the scene.  

\noindent \textbf{Relation Plane Change} - Assigns an object to another plane. 

\noindent \textbf{Resample} - Regenerates an object with new parameters.

\noindent \textbf{Reinitialize Pose} - Samples a new random pose for an object.

\noindent \textbf{Translate} - Translates the object within its DoF plane.  

\noindent \textbf{Rotate} - Rotates the object around its DoF axis.

We observe that not all moves are equally significant at each point in the optimization. In an empty scene, object addition and relation change allow for higher loss reduction, whereas in a cluttered scene, continuous object movement allows for higher loss reduction. Thus, we provide a schedule for moves so that probabilities for discrete moves decay gradually and probabilities for continuous moves increase.  

The objects in an indoor scene are interdependent on each other, which makes it unfeasible to optimize over all of them simultaneously. However, they usually depend on each other hierarchically (e.g. a cup is on the table, which is on the floor). To exploit this hierarchy, we divide our optimization into three stages: large object optimization, medium object optimization, and small object optimization. 

Each object is constrained in its movement due to the constraints specified by the user and the discrete moves proposed by the solver. For instance, a bookshelf that is stable against a wall is only allowed to move along the 1D line between the wall and the floor. Consequently, an object's degrees of freedom (DoFs) for rotation and translation are determined based on its relations to other objects. When the solver samples a continuous move, it restricts the object's motion to these DoFs. When the solver samples a discrete move, it places the object in the constrained subspace. 
\vspace{-1em}
\paragraph{Floorplan-specific solver and constraints}
Our floorplan generator creates realistic full-house room meshes, as shown in Fig.~\ref{fig:floor_plan}. First, we procedurally generate a room adjacency graph specifying the number, type, and connectedness of individual rooms required in the floor plan. This graph is produced by inference on a probabilistic context-free grammar on room types, or can be wholly or partially derived from user input. We define our objective function as a weighted combination of the terms below, and optimize it using simulated annealing subject to constraints from the room adjacency graph. See Appendix \ref{sec:floor_plan_solver} for full definitions.

\begin{tabular}{ll}
\labelitemi\  Shortest path to entrance &\labelitemi\ Typical room area\\
\labelitemi\  Room aspect ratio &\labelitemi\ Room convexity\\
\labelitemi\  Room wall conciseness &\labelitemi\ Functional room area \\
\labelitemi\  Room collinearity &\labelitemi\ Narrow passages \\
\labelitemi\  Exterior length by room &\labelitemi\ Exterior corners by room \\
\labelitemi\  Staircase occupancies &\labelitemi\ Staircase IOU with rooms \\
\end{tabular}

 We initialize our floorplan solver by generating a random house outline, and subdividing it using a Mondrian Process \citep{Roy2008TheMP} until it produces sufficient spaces for each room. We extrude a wall segment inwards or outwards at each step, or swap the assignment subject to the room adjacency graph. Either action will lead to a change in loss, which we convert to an acceptance probability using Metropolis-Hasting as in Sec \ref{sec:solver}. Once solving is complete, we add floor, wall, and ceiling materials, doors, windows, and staircases, all subject to constraints based on room type and adjacency. Fig. \ref{fig:room_final} shows the solutions to room arrangements with objects placed inside. 

\subsection{Data Export}
\begin{figure}
    \centering
    \resizebox{\linewidth}{!}{%
        \includegraphics[width=0.5\textwidth]{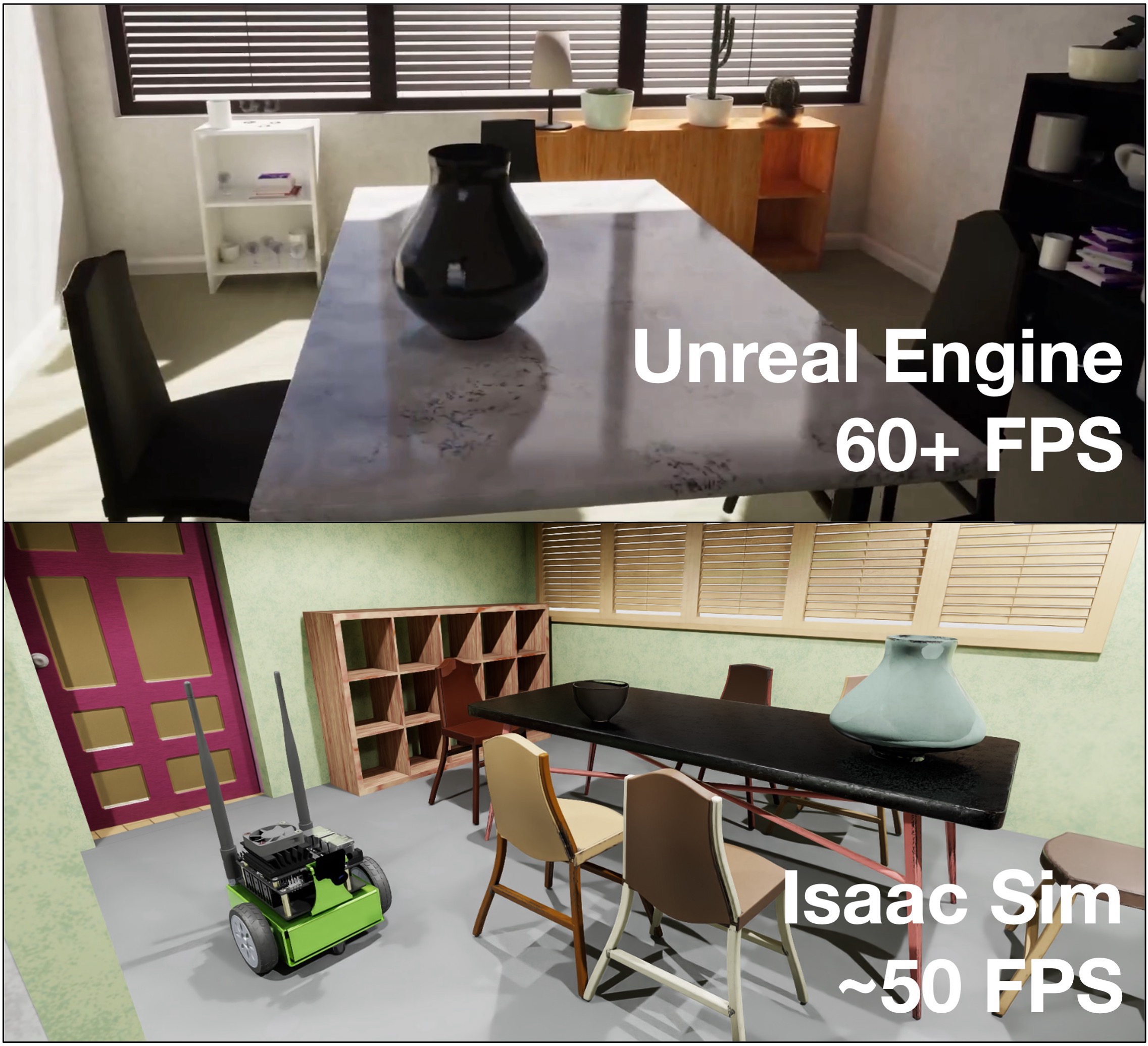}%
    }
    \vspace{-.5em}
    \caption{Two scenes imported into Unreal Engine 5 (above) and Omniverse Isaac Sim (below). Unreal Engine runs at 60 FPS, and Isaac Sim, with physics simulation enabled, runs at 50 FPS, both on RTX 4090s.}
    \label{fig:sim}
    \vspace{-1em}
\end{figure}

\begin{table}[h]
\begin{center}
    \resizebox{\columnwidth}{!}{
        \begin{tabular}{l|l|cccccc}
        \toprule
        \multicolumn{2}{c}{} & \multicolumn{2}{|c}{All Image} & \multicolumn{2}{c}{Shadow Region} & \multicolumn{2}{c}{Non-Sh Region}	\\ 
        \midrule
        Test Set & Model & PSNR$\uparrow$ & RMSE$\downarrow$ & PSNR$\uparrow$ & RMSE$\downarrow$ & PSNR$\uparrow$ & RMSE$\downarrow$\\
        \midrule
        \multirow{3}{*}{ISTD \cite{istd}} & R & \textbf{31.96} & \textbf{4.27} & \textbf{38.04} & \textbf{6.58} & \textbf{34.13} & \textbf{3.85}\\
        & R+S & 31.72 & 4.30 & 37.41 & 7.06 & 33.80 & 3.85\\
        \midrule
        \multirow{3}{*}{SRD \cite{srd}} & R & 22.83 & 10.84 & 25.59 & 18.75 & 27.93 & 7.68\\
        & R+S & \textbf{24.56} & \textbf{9.54} & \textbf{27.39} & \textbf{16.63} & \textbf{29.37} & \textbf{6.67}\\
        \bottomrule
        \end{tabular}
    }
    \caption{Shadow removal task quantitative performance on ISTD, ISTD+, and SRD dataset across the three variations of the model. }
    \label{tab:shadow_tab}
    \end{center}
    \vspace{-1.5em}
\end{table}

\begin{figure}
  \begin{center}
  \resizebox{\linewidth}{!}{
    \includegraphics[width=\textwidth]{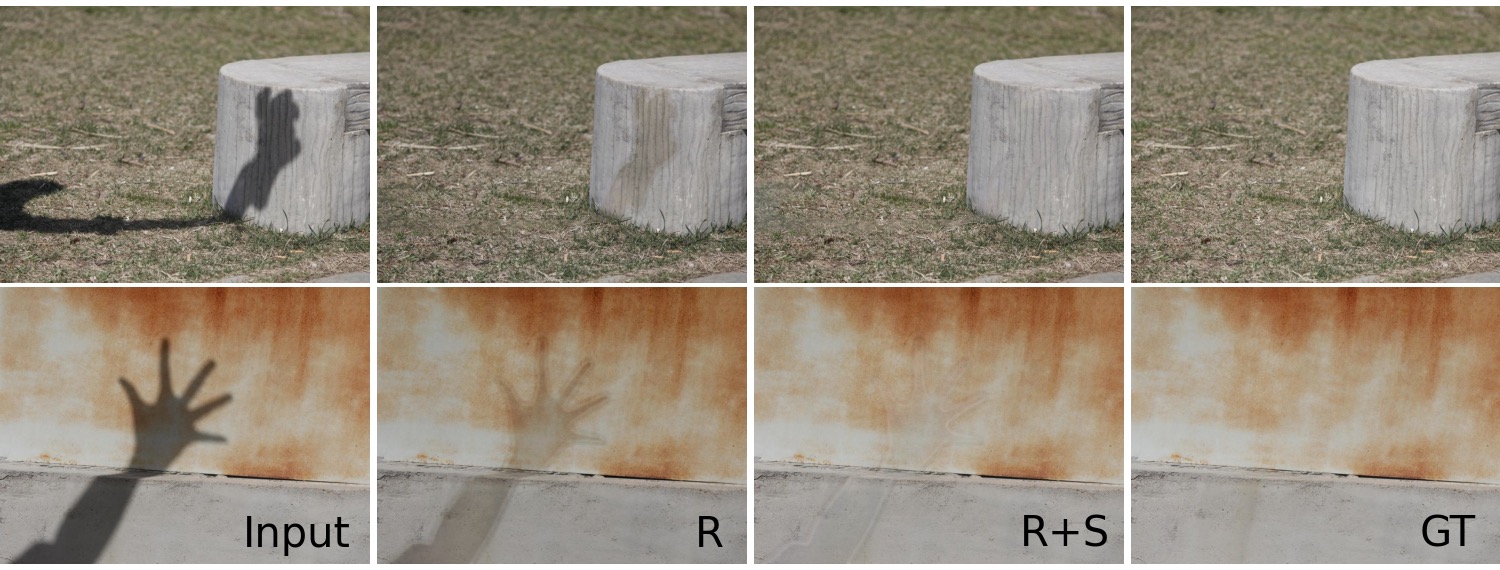}
    }
  \end{center}
  \vspace{-1.5em}
  \caption{Qualitative zero-shot results on SRD \cite{srd} test dataset.}
  \label{fig:shadow_qualitative}
\end{figure}

We develop a one-click tool to export assets from Infinigen Indoors to real-time simulators, using Universal Scene Description (USD) or other formats. As seen in Fig.~\ref{fig:sim}, indoor scenes can be exported to Omniverse Isaac Sim and Unreal Engine 5 and can be run at interactive frame rates. This exporting capability allows Infinigen Indoors to help train embodied agents in virtual environments. 

Infinigen Indoors uses Blender's procedural material system, which is by default not portable to other simulators or scene editors. To resolve this, we provide tools to automatically post-process and UV-map entire indoor scenes, and use texture baking to create standard texture maps for material color, roughness, metallicity and more. We also provide export code to convert single objects to textured OBJ, FBX or STL meshes, and automatically generate collision and articulation information as Universal Robot Description Format (URDF) files. 

\section{Experiments}

\begin{figure}
  \begin{center}
  \resizebox{\linewidth}{!}{
    \setlength{\tabcolsep}{0.6pt}
    \begin{tabular}{ccccc}
      Image &
      GT &
      Infinigen-Nature~\cite{infinigen2023infinite} &
      Ours & 
 \\
    \galleryRowCompare{figures/occlusion_qual_results}{media_images_images_0.jpg}{media_images_true_0_gt.jpg}{media_images_pred_0_out.jpg}{media_images_pred_0_in.jpg}
    \galleryRowCompare{figures/occlusion_qual_results}{media_images_images_1.jpg}{media_images_true_1_gt.jpg}{media_images_pred_1_out.jpg}{media_images_pred_1_in.jpg}
    \end{tabular}
}
  \end{center}
  \vspace{-1.5em}
  \caption{Qualitative results on synthetic artistic scenes~\cite{gumroad}.}
  \label{fig:occlusion_qual}
\vspace{-1.5em}
\end{figure}

\begin{figure*}[th!]
    \centering
    \includegraphics[width=\textwidth]{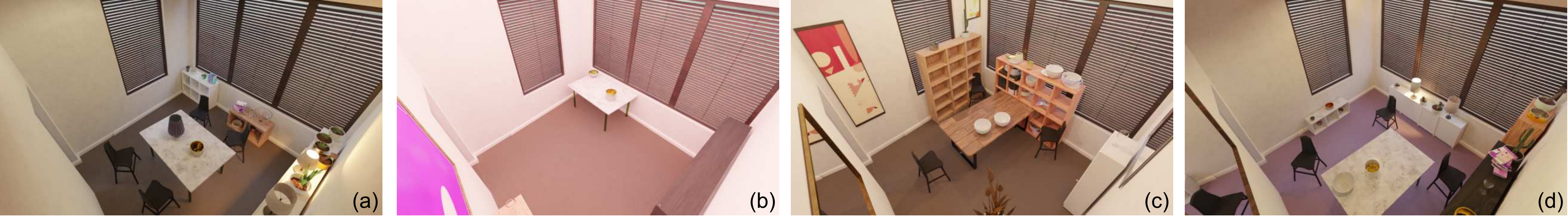}
    \caption{Qualitative Ablation. From left to right, we show scenes generated by our system with 10K solving steps (a), 1K solving steps (b), with collision checking removed (c) and with symmetry terms disabled (d)}
    \label{fig:qual_ablation}
\end{figure*}

\subsection{Solver Performance}
To efficiently solve large numbers of constraints in cluttered scenes, we optimized our solver with various features for faster convergence. Plane hashing enables faster access to bounding planes during discrete optimization. BVH caching enables faster mesh distance and collision calculations by reusing Bounding-Volume-Hierarchies except when mutated by a state update. Evaluation caching maintains a cache of results in the evaluation graph. Move filtering narrows down the search space in continuous optimization by selectively pruning candidates to those that can reduce the loss. Placeholder optimization only generates full meshes when other objects are assigned to them; otherwise it keeps bounding boxes. 

To analyze the importance of these features, we conducted an ablation in Tab. \ref{tab:solver_perf} and Fig. \ref{fig:qual_ablation}, which shows solver performance with each feature removed. All results are averaged over 20 random scenes with 5k solver steps. Our full system provides a $\approx 3$x speedup compared to the non-optimized version. Most of the performance gains come from BVH caching and Plane Hashing. We observe that discrete changes such as pose re-initialization, relation changes, and object resampling are necessary for compelling visuals, but decrease the quantitative score and increase the runtime. When we run our fully optimized system as long as the non-optimized version we get a $28\%$ score increase. 
\begin{table}[h]
\centering
    \resizebox{\columnwidth}{!}{
    \begin{tabular}{l|c|l|l}
    \toprule
     Method & Runtime $\downarrow$ & Avg. Score (s) $\uparrow$ & \#Objects $\uparrow$ \\
     \midrule
     \textbf{Full System} & 2280.68 & 80.84 & 28.94 \\
     w/o Discrete change & \textbf{1872.60} & 86.20 & 30.82\\
     w/o Eval. cache & 2242.79  & 80.84 & 28.94\\
     w/o Move Filter & 2633.95 & 84.75 & 45.06 \\
     w/o Placeholder optim. & 2665.18  & 80.46 & 27.71 \\
     w/o Plane Hashing & 3522.86 & 92.65 & 34.12 \\
     w/o BVH cache & 4740.48 & 80.84 & 29.06 \\
          \textbf{Full System 100 min.} & 6080.84 & \textbf{114.72} & \textbf{49.65}\\
     w/o Optimization & 6308.92 & 89.43 & 47.76 \\
     \bottomrule
    \end{tabular}
}
\caption{Ablation of solver performance optimizations.}
\vspace{-3mm}
\label{tab:solver_perf}
\end{table}

\noindent \textbf{Perceptual Study} We performed a crowdsourced human evaluation of our scenes and layouts, following metrics from ATISS \cite{atiss}. Table \ref{tab:ustudy_res} shows that the subjects preferred Infinigen Indoors over \cite{Deitke2022ProcTHORLE, atiss, scene_former, ritchie_fast_2018} in terms of both realism, layout realism, and the lack of errors, although we recognize ``realism'' may also have been influenced by asset and lighting quality. See section \ref{sec:perceptual} for more details.

\subsection{Shadow Removal}

To demonstrate its flexibility in data generation, we used Infinigen-Indoors to create a dataset consisting of 2k image pairs of shadow and shadow-free variants. These pairs were generated by toggling the shadow property of lighting within Blender. For each pair, shadow masks were produced using Otsu's thresholding \cite{otsu} method. We use ShadowFormer \cite{shadowFormer} model for the experiments and consider two variations: only trained on ISTD \cite{istd} real dataset (R), and trained on combination of ISTD and 2k \projectname{} synthetic dataset (R+S). The results are shown in Tab. \ref{tab:shadow_tab}. While using synthetic data leads to slightly worse performance on the ISTD dataset, the zero-shot application on the SRD \cite{srd} dataset shows clear improvement for generalization to new test datasets. Qualitative results are shown in Fig.~\ref{fig:shadow_qualitative}.

\subsection{Occlusion Boundary Estimation}

\begin{table}[h]
\centering
\begin{tabular}{l|c|c|c} \toprule
 Training Dataset & ODS & OIS & mAP \\
 \midrule
 Infinigen-Nature~\cite{infinigen2023infinite} & $14.38$ & $19.43$ &$10.80$\\
 Hypersim~\cite{Roberts_2021_ICCV} & $26.02$ & $19.44$ & $15.69$\\
 \projectname{} (Ours) & $\mathbf{29.47}$& $\mathbf{30.29}$ & $\mathbf{19.09}$\\
 \bottomrule
\end{tabular}
\caption{Occlusion boundary quantitative results on a curated test set of photorealistic artist-designed synthetic 3D scenes for architecture visualization~\cite{gumroad}. }
\label{tab:occlusion}
\end{table}

To validate the effectiveness of Infinigen Indoors, we also evaluate on the task of occlusion boundary estimation, a task with limited available data. We produce 1464 images annotated with ground truth. We train three U-Net~\cite{ronneberger2015u} models from scratch separately on these images, on images generated from Infinigen~\cite{infinigen2023infinite} and Hypersim~\cite{Roberts_2021_ICCV}. We then compare their performance on a curated test set of photorealistic artist-designed synthetic 3D scenes for architecture visualization~\cite{gumroad}, since no existing photorealistic indoor dataset provides such annotations. See Appendix \ref{sec:app_exp} for more details. 
 
 We report the following three metrics~\cite{wang2020occlusion, weinzaepfel2015learning, wang2019doobnet}: (i) \textit{optimal dataset scale F-score (ODS)}, representing the best F-score achieved on the dataset using a uniform threshold across all test images; (ii) \textit{optimal image scale F-score (OIS)} indicating the cumulative F-score on the dataset obtained with thresholds dependent on individual images; and (iii) \textit{mean average precision (mAP)} denoting the mean precision across the complete recall range.

As we can see in Tab. \ref{tab:occlusion}, our \projectname{}-trained model generalizes better. The model achieves higher performance across all metrics. These findings underscore the usefulness of \projectname{} as a valuable training resource. Qualitative results are depicted in Fig.~\ref{fig:occlusion_qual}.

\section{Contributions \& Acknowledgements}
Alexander Raistrick, Lingjie Mei and Karhan Kayan contributed equally and are ordered randomly. Each has the right to list their name first in their CV. Alexander Raistrick performed team coordination and developed the constraint language and object graph solver. Lingjie Mei developed the room solver and many procedural assets. Karhan Kayan developed the geometric constraints and object pose solver. David Yan developed the scene exporter and other utilities. Yiming Zuo, Beining Han, Hongyu Wen, Meenal Parakh, Stamatis Alexandropoulos, Zeyu Ma and Lahav Lipson developed procedural assets and utilities. Jia Deng conceptualized and led the project, and set directions. This work was partially supported by the National Science Foundation and Amazon.

{
    \small
    \bibliographystyle{ieeenat_fullname}
    \bibliography{main}
}

\newpage
\appendix
\section*{Appendix}

\begin{figure*}[ht]
\centering
\includegraphics[width=0.9\linewidth]{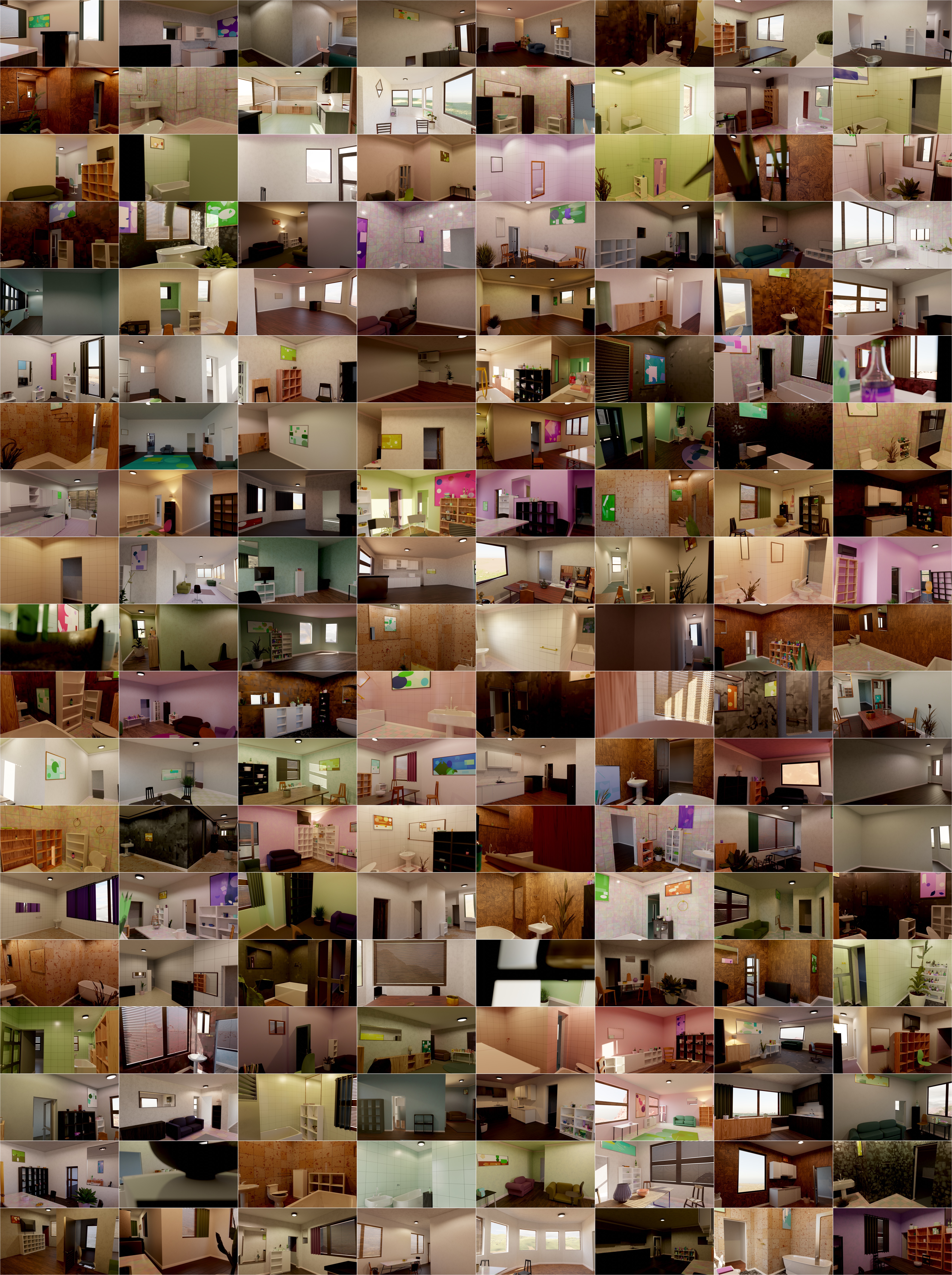}
 \caption{Random, non-cherry-picked sample of procedurally generated residential homes (Part 1 of 2)} 
\label{fig:rand_1}
\end{figure*}

\begin{figure*}[ht]
\centering
\includegraphics[width=0.9\linewidth]{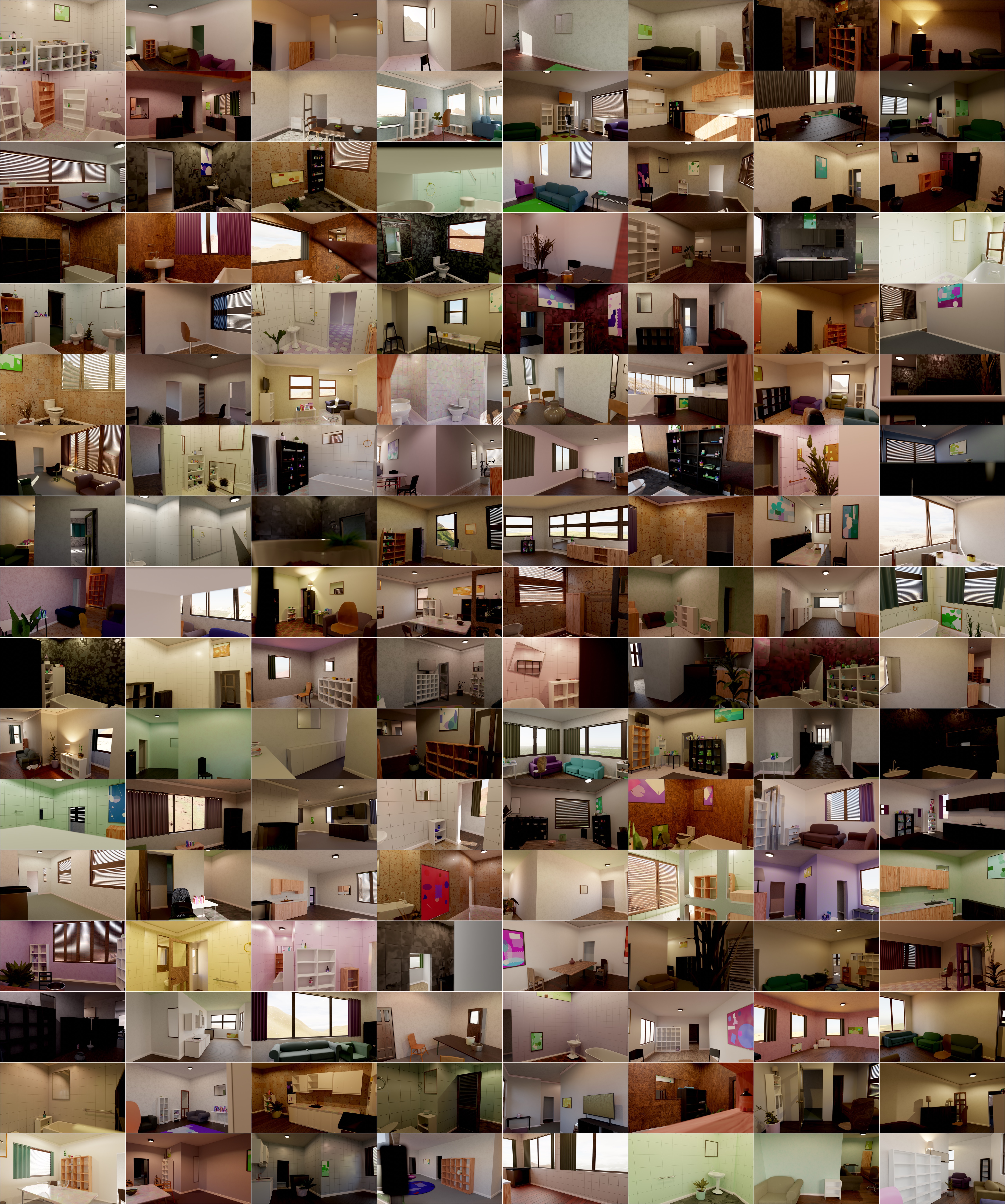}
 \caption{Random, non-cherry-picked sample of procedurally generated residential homes (Part 2 of 2)\vspace{2cm}} 
\label{fig:rand_2}
\end{figure*}

\begin{figure*}
\includegraphics[width=\linewidth]{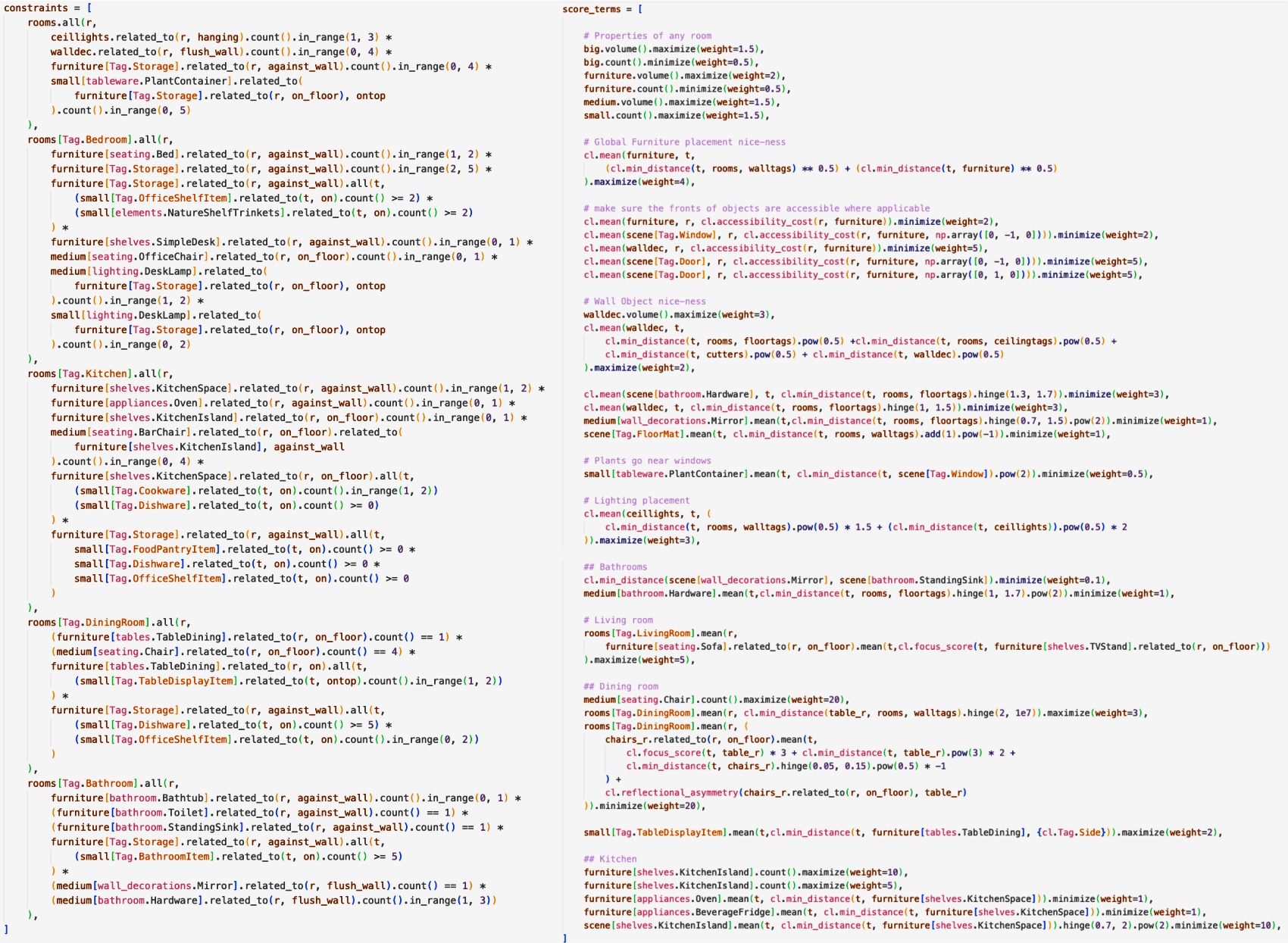}
 \caption{Constraint Program for whole residential homes. Left shows hard constraints, right shows continous objective scores. Our system flexibly composes API calls to apply any constraint to any class of object describable within the scene.} 
\label{fig:rooms_code}
\end{figure*}

\section{Constraint Specification API}

\subsection{API Description}
\label{sec:api_desc}

\begin{table*}[t]
    \centering
    \resizebox{\linewidth}{!}{%
    \begin{tabular}{l|c|l} 
        \toprule
            Optimization objectives based on... & Our API Function & Example Usage (Described in Natural Language) \\
        \midrule
            Objects with (abstract) semantics & \texttt{with\_semantics} a.k.a. \texttt{$[\hspace{2mm}]$} & Scope a constraint to a hierarchical class e.g. shelves, storage, all furniture, or all objects\\
            Objects related to other objects & \texttt{related\_to\(\)} & Cooking pots go in the center when on tables, but can go anywhere on a countertop. \\
            Objects on arbitrary surfaces & \texttt{SupportedBy} & Multi-story homes, decorations on shelves, countertops, fridges \\
            Whether objects overhang & \texttt{StableAgainst} & Paintings cant overhang walls \\
            Non-convex object shapes & Yes, procedural placeholders & Objects go inside shelves, chairs tuck under table \\
        \midrule
            Variable quantity of objects & \texttt{count} & Allow between 0 and 3 sofas in a living-room, but as many as possible \\
            Size of objects & \texttt{area}, \texttt{volume} & Generate the biggest possible TV \& Sofa that fits well\\ 
            Pair-wise distances & \texttt{min\_distance} & Place dining tables \& ceiling lights far from walls \\
            Pair-wise angle difference & \texttt{angle\_alignment} & Align tables to parallel to the nearest wall \\ 
            Symmetry around an object & \texttt{rotational\_asymmetry} & Chairs should be rotationally symmetric when placed around circular tables \\
            Symmetry across a plane & \texttt{reflection\_asymmetry} & Bed-side tables should be symmetric on either side of a bed\\ 
            Objects facing other objects & \texttt{focus\_score} & Sofas should face TVs or paintings \\ 
            Object accessibility & \texttt{accessibility\_cost}  & Leave space in front of sofas / appliances \\ 
            Empty space on a surface & \texttt{freespace\_2d} & Leave some space leftover in room / on countertop \\ 
        \midrule
            Arbitrary arithmetic / nonlinearities & \texttt{+ - * / pow hinge}  & Encourage certain ratio of ceiling-lights to room-area \\
            Boolean comparisons / logic & \texttt{== < <= and in\_range} & Ensure there are 2 to 6 chairs for every table \\
        \midrule
            \textit{every} object must satisfy a predicate & \texttt{all} & Every bookcase must have $\geq 10$ books \\
            sum/mean across specific objects & \texttt{mean sum} & Compute average distance to wall over many objects, rather than minimum \\ 
        \bottomrule
    \end{tabular}}
    \caption{Capabilities included in our API. Please see Sec. \ref{sec:rand_sample} and \ref{sec:warehouses} for example programs, and surrounding text for full descriptions. For our API, functions can be composed arbitrarily, e.g. \texttt{scene.with\_semantics(...).related\_to(...).count().hinge(...)} to create a nonlinear objective w.r.t. number of objects in a certain context.}
\end{table*}

\paragraph{\texttt{with\_semantics}} (also notated as operator square-brackets for brevity) extracts the subset of a set of objects that satisfies a semantic predicate, e.g. "extract the subset of rooms which are dining-rooms" or "extract the subset of furniture objects which are shelving". The hierarchy of these predicates is defined by asset creators, or can be reconfigured by constraint program writers if they intend to use an object for an unusual purpose. Using hierarchical classes for this filtering operation allows every constraint to apply to the most broad class of objects possible, to avoid rewriting or restating constraints for objects that fulfill a similar function.

\paragraph{\texttt{related\_to}} extracts the subset of a set of objects related to any member of a second set of objects via some relation. The exact way in which the objects are related is user-configurable by passing in any parameterized \texttt{Relation} object from the options below (StableAgainst, SupportedBy), which represents a predicate that can be True or False of any pair of objects. 

By combining \texttt{related\_to} with other filtering operations, the user can express constraints on arbitrarily complex contexts such as "maximize the number of dining chairs against dining tables inside of rooms adjacent to kitchens", to encourage plenty of seating near food preparation areas, etc. 

\paragraph{\texttt{scene}} retrieves the set of all objects currently in the scene. All constraint programs are ultimately functions of the current scene state, so this node serves as the leaf node of all constraint program expressions (besides numeric constants). Users rarely place constraints on \texttt{scene} directly. Instead, we expect the user first to take subsets via the operations above.

\paragraph{\texttt{StableAgainst}}
specifies a relation using a child object's planar surface, a parent object's planar surface, and a margin between the surfaces. It checks that the child's surface is parallel to the parent's, the child is not overhanging, and the child's surface is exactly at the specified margin. A concrete example would be specifying the sofa as stable against the floor with zero margin and stable against the wall with a 10cm margin. Alternatively, specifying a painting to be stable against the wall ensures that the painting does not overhang across the edge of the wall. 

\paragraph{\texttt{SupportedBy}}
specifies a relation using a child object's planar surface and a parent object's planar surface. It means that the child object would not fall over from the parent object. More precisely, the surfaces are parallel against each other with zero margins, and the centroid of the child object is contained within the convex hull of the intersection between the child and the parent object. The last condition is to ensure zero torque by gravity. An example use case is a coffee cup teetering on the table's edge. In this case, the cup is supported by the table, but it is not stable against it since it is overhanging. 

\paragraph{\texttt{count}} returns the cardinality of a set of objects in the scene. 

\paragraph{\texttt{area, volume}} returns the total area or volume of the bounding boxes of objects in a set. We use bounding boxes to avoid expensive calculations to compute the exact volume of each mesh, and we find this serves as a suitable proxy to incentivize larger assets. Area always takes over the two largest axes of an object and is usually used for 2D objects like paintings or rugs.

\paragraph{\texttt{min\_distance}}
calculates the minimum distance between two sets of tagged objects. For instance, the minimum distance between the walls and the back of the couch. The minimum distance is defined as the distance between the closest two points on the two sub-meshes identified by the tags.

\paragraph{\texttt{angle\_alignment\_cost}}
quantifies how far a group of objects are from being angle aligned to a reference object on the XY plane. The cost is calculated as
\[ \sum_i \frac{1 -\cos \theta_i}{2}\]
where $\theta_i$ is the angular difference between the front-facing normal of object $i$ and the inward normal of the closest edge of the reference object. The contribution of each object is in the $[0,1]$ range. 

An example use case is minimizing the angle alignment cost between chairs and tables to make the chairs face the table. Another example is using an alignment score to align furniture to the walls in order to give the arrangement a more grid-like shape. 

\paragraph{\texttt{rotation\_asymmetry}}
gives a continuous characterization of the rotational asymmetry of a set of objects based on \cite{zabrodsky_continuous_1992, frey_method_2007}. It measures the deviation of the set of objects from a regular polygon with perfectly rotationally symmetric orientations. From another perspective, it measures the rotational asymmetry of a set of point-vector pairs. The score consists of two parts and is calculated as 
\[\text{score} = \frac{\text{location asymmetry} + \text{orientation asymmetry}}{2}.\]
Suppose the location of the $i$th object is given by $\vec{x}_i$ and there are $n$ objects. The location asymmetry is calculated as follows:
\begin{itemize}
    \item Let $\vec{p}_i = \vec{x}_i - c$ where $c$ is the centroid of the objects.  
    \item Rotate all $\vec{p}_i$ so that $\vec{p}_1$ is aligned with the axis.
    \item Normalize $\vec{p}_i$ by dividing by $\max_i ||\vec{p}_i||$.
    \item Let $\vec{f}_{i}$ be vector $\vec{p}_i$ rotated by $-2i \pi/n$.
    \item Compute $\vec{q}$ as the average of $\vec{f}_{i}$.
    \item Let $\vec{w}_i$ be vector $\vec{q}$ rotated by $2i \pi/n$. 
    \item Then, we have $\text{location asymmetry} = \frac{1}{n} \sum ||\vec{w}_i - \vec{p}_i||^2$.
\end{itemize}
The orientation asymmetry score follows the same steps as the location asymmetry, but with $\vec{p}_i$ replaced by the frontal plane normal of the object $i$. 

As an example, rotational asymmetry score can be used to encourage tableware being rotationally symmetric on the table not only with respect to their location but also their orientation. It can also be used to make chairs rotationally symmetric around the table. 

\paragraph{\texttt{reflection\_asymmetry}}
calculates a continuous reflectional asymmetry score for a set of objects relative to a reference object. This score quantifies the deviation of objects from mirror symmetry. The process involves reflecting each object across a plane and then comparing the original and reflected objects. From another perspective, it quantifies the mirror asymmetry of a set of point-vector pairs. The asymmetry score is computed as follows:
\begin{itemize}
\item \textbf{Determine Reflection Plane:} Identify the plane of reflection, which can be any of the median planes of the bounding box of the reference object.
\item \textbf{Reflect Objects:} The objects \(O_i\) are represented by $(\vec{p}_i, q_i)$ where $\vec{p}_i$ is the object's location and $q_i$ is the object's orientation. Each object \(O_i\) is reflected across the plane to obtain its mirror image \(O'_i\). The reflection of a point \(\vec{p}\) is given by \(\vec{p'} = \vec{p} - 2(\vec{p} \cdot \vec{N}_{plane})\vec{N}_{plane}\). The reflection of an axis-angle represented orientation $\theta \vec{e}$ is given by $\theta' \vec{e'}$ where $\vec{e'} = \vec{e} - 2(\vec{e} \cdot \vec{N}_{plane})\vec{N}_{plane}$ and $\theta' = -\theta$. 

\item \textbf{Bipartite Matching:} A cost-minimizing bipartite matching is performed between the set of original objects \(\{O_i\}\) and their reflected counterparts \(\{O'_i\}\) to find the optimal pairings based on a cost matrix derived from positional and angular deviations. We use a modified Jonker-Volgenant algorithm for this step \cite{jonker_shortest_1987, crouse_implementing_2016}. 

\item \textbf{Calculate Deviations:}
    \begin{itemize}
        \item \textbf{Positional Deviation:} For each paired object \((O_i, O'_i) = ((\vec{p}_i, q_i),(\vec{p'}_i, q'_i))\), calculate the Euclidean distance \(D_{pos} = ||\vec{p}_i - \vec{p'}_i||\).
        \item \textbf{Angular Deviation:} Calculate the angular difference \(D_{ang} = 2 \arccos(|q_i \cdot q'_i|)\), where \(q_i\) and \(q'_i\) are the quaternion representations of the paired objects' orientations.
    \end{itemize}

\item \textbf{Weight Deviations:} Each deviation is weighted by a factor \(V(O_i)\), which is the volume of the bounding box of $O_i$. The weighted deviation for each object pair is \(D_{dev}(O_i) = V(O_i) \times (D_{pos} + D_{ang})\).

\item \textbf{Normalization:} The total deviation is normalized by a factor \(\alpha\), which is the average distance between objects: \(\alpha = \frac{1}{N(N-1)}\sum_{i \neq j} ||\vec{p}_i - \vec{p'}_i||\), where \(N\) is the number of objects.

\item \textbf{Compute Asymmetry Score:} The reflectional asymmetry score is derived as 
\[\text{Score} = 1 - \frac{1}{1 + \sum_i D_{dev}(O_i) / \alpha}\]
\end{itemize}

This reflection score is useful in contexts such as encouraging chairs to be symmetric around a long rectangular table, or encouraging furniture to have mirror symmetry for visual appeal, or encouraging paintings to be symmetrical across the room.

\paragraph{\texttt{accessibility\_cost}}
computes how much a set of objects $B$ block access to a set of objects $A$. We offer two versions. In the fast version, the function selects the closest object in \( B \) to each object in \( A \) based on the centroid distance. In the slow version, it finds the closest point on any mesh in \( B \) to each mesh in \( A \). The mathematical formulation can be described as follows:

\noindent We first take the projection of \( a \)'s centroid onto its specified plane (frontal plane by default) by 
\[
\vec{a}_{\text{proj}} = \vec{a}_c - \left( (\vec{a}_c - \vec{f}_p) \cdot \vec{n}_a \right) \vec{n}_a
\]
where \( \vec{a}_c \) is the centroid of object \( a \), \( \vec{f}_p \) is a point on the specified plane, and \( \vec{n}_a \) is the normal vector of the specified plane.

    \noindent For a given object $a \in A$, we define \( \vec{b}(a) \) and \( \vec{b}_{\text{closest\_pt}} \). The fast version defines
     \begin{flalign*}
     &\vec{b}(a) = \argmin_{b \in B} \|\vec{b}_c - \vec{a}_{\text{proj}}\| & \\
     &\vec{b}_{\text{closest\_pt}} = \text{Centroid of the selected } \vec{b}(a) &
     \end{flalign*}
The slow version defines
     \begin{flalign*}
     &\vec{b}(a) = \text{Object in } B \text{ with the point closest to mesh } a & \\
     &\vec{b}_{\text{closest\_pt}} = \text{Point on mesh } \vec{b}(a) \text{ closest to mesh } a &
     \end{flalign*}

\noindent For both fast and slow versions, the accessibility cost is calculated as
\[
\text{cost} = \sum_{a \in A} \frac{(\vec{b}_{\text{closest\_pt}} - \vec{a}_{\text{proj}}) \cdot \vec{n}_a}{\|\vec{b}_{\text{closest\_pt}} - \vec{a}_{\text{proj}}\|^2} \times \|\vec{b}(a)_d\|
\]
where \( \vec{b}(a)_d \) is the diagonal vector of the bounding box of the chosen object \( \vec{b}(a) \). We note that the accessibility cost increases as the blocking object gets larger, as the blocking objects get closer, and as the blocking object is more in front of the specified plane. 

An example usage of accessibility cost is when we want to penalize objects being directly in front of TVs, paintings, or closets. 

\paragraph{\texttt{focus\_score}}
encourages focusing a set of objects $A$ on an object $b$. It is calculated as 
\[ \sum_{a \in A} \frac{1 - \vec{n}_a \cdot (\vec{b}_c - \vec{a}_c)}{2||\vec{b}_c - \vec{a}_c||}\]
where $\vec{n}_a$ is the front facing normal of object $a$. The vectors $\vec{a}_c, \vec{b}_c$ denote the centroids of $a$ and $b$ respectively. The contribution of each object is in the $[0,1]$ range. 

An example use case of focus score is focusing the sofas on the TV to encourage a more realistic layout. Another example use case is focusing a set of seats on a round table. 

\paragraph{\texttt{freespace\_2d}}
returns the amount of 2D free space available on a set of objects $A$ after accounting for the space occupied by objects $B$. It is calculated as 
\[ \sum_{a \in A} \mu(\mathrm{proj}(a)) - \sum_{b \in B} \mu(\mathrm{proj}(b))\]
where $\mathrm{proj}$ is the projection to the XY plane and $\mu$ gives the area of a 2D shape. An example use case is minimizing the free space on a table to encourage placing more objects on the table, or maximizing the free space in a living room to make it less cluttered. 

\paragraph{Arithmetic / non-linearities} provide basic scalar arithmetic and an implementation of the standard hinge loss function, all computed using the standard Python definitions. The exact set of mathematical operators provided here is not critical; our system treats scalar losses as a black box, so any arbitrary Python math expressions are acceptable. 

\paragraph{Boolean comparisons} allow equality or inequality checking between values, usually for creating hard constraints on cardinalities or distances. When used to check the size of a set, our system will use the constraint statement to inform what Addition moves are proposed as explained in ``Cardinality Bounding".

\paragraph{\texttt{all}} provides control flow logic akin to the ``forall'' $\forall$ symbol as used in formal proofs. It is commonly used in constructing soft/hard constraints. We avoid allowing arbitrary Python control flow (for loops, if statements, etc.) as it makes symbolic reasoning difficult by restricting the user to symbolic expressions. This design decision is similar to other compute graph programming frameworks (e.g. Tensorflow \cite{tensorflow2015-whitepaper}, CVXPY \cite{diamond2016cvxpy}). For example, by forcing the user to use a symbolic 'all' statement rather than a for loop, we can make inferences such as "if all chairs go near tables, and there must be at least one chair, then there must be at least one table", which allows the user to write higher level and fewer constraints, with the system deducing all logical consequences.

Forall statements take as input a loop variable name, and a constraint program that contains the loop variable as a leaf node at one or more locations. During execution, the child constraint program is substituted with the real values of the loop variable and evaluated to obtain the various results. 

\paragraph{\texttt{mean, sum}} compute the standard scalar mean and sum operations, using similar control flow logic and evaluation substitution mechanisms as \texttt{all} as described above.

\section{Extended Random Sample \& Constraint Code for Residential Scenes}
\label{sec:rand_sample}

Please inspect Fig. \ref{fig:rand_1} and Fig. \ref{fig:rand_2} for an extended random sample of our main residential home generator (as shown in Fig. \ref{fig:teaser}).

These images were derived from the constraint code designed for residential homes, shown in Figure \ref{fig:rooms_code}. This code uses a total of 105 soft and hard constraints, with 19 for dining rooms, 14 for living rooms, 9 for bathrooms, 18 for kitchens, 16 for warehouses, and 30 which apply abstractly to all rooms. These constraints are used to cover object assignments (object A goes on object B), ratios (numbers of chairs per table, objects per shelf), stability (TV placed against the wall; objects don't overhang unsupported), distance (plants placed near window), and more. 

\begin{figure*}
\includegraphics[width=\linewidth]{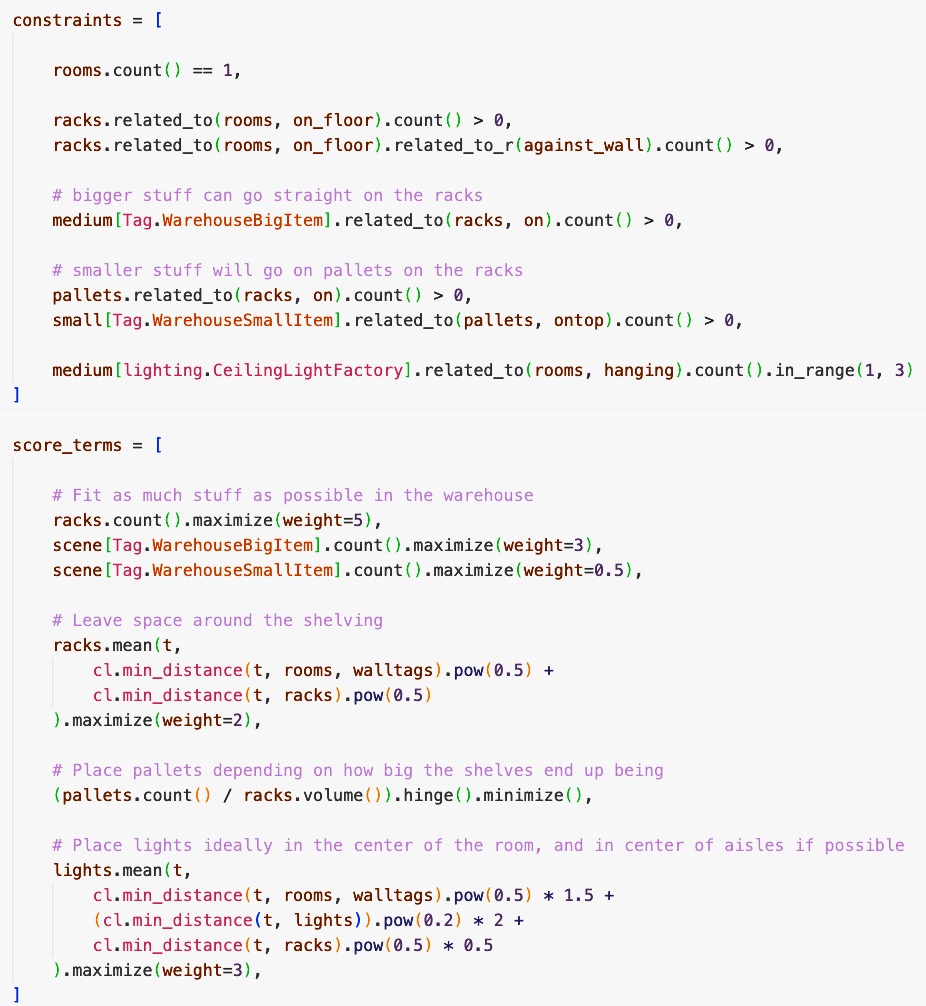}
 \caption{Constraint program for warehouse scenes. In only a few high-level statements, we specify the hierarchy of allowed objects, and competing placement objectives that give rise to an appropriate shelf and object layout for any warehouse scene.} 
\label{fig:warehouse_code}
\end{figure*}

\section{Extension to warehouse scenes}
\label{sec:warehouses}

\begin{figure*}
\includegraphics[width=\linewidth]{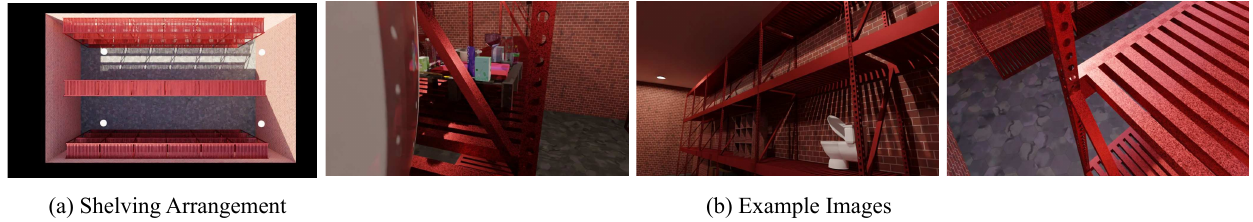}
 \caption{Warehouse scene arrangement (left) and example first-person images (right). Using only a few high-level objectives, we extend our existing placement system and existing furniture generators to create a hardware-store-like environment.}
\label{fig:warehouse_img}
\end{figure*}

To show the generality of our solving system, we implemented a simple constraint program that uses existing language features to specify the high-level objectives of a warehouse environment, with furniture on shelves and smaller items on wooden pallets. See Fig. \ref{fig:warehouse_code} for the full program. Various further extensions are possible, for example indicating a preference for larger objects to be placed lower or higher on the shelves, or certain objects to be placed near the front of the warehouse / store. We show example images in Fig. \ref{fig:warehouse_img}, as well as a topdown view showing only the shelving and lighting layout. 

\section{Floorplan Solver Details}
\label{sec:floor_plan_solver}
\subsection{Floor plan graph generation}
We generate floor plans that have 1 to 3 floors. For each floor, we generate a floor plan graph where individual nodes represent a room with a certain type, and each edge represents the connectedness of two rooms it is linked to. We support rooms of the following type: {\it kitchen}, {\it bedroom}, {\it living-room}, {\it closet}, {\it hallway}, {\it bathroom}, {\it garage}, {\it balcony}, {\it dining-room}, {\it utility}, {\it staircase}. {\it hallway} can mean any corridor or passage between rooms, and {\it staircase} means the room or space where one can find the stairs. 

The graph is generated by a Probabilistic Context-free Grammar(PCFG), where the graph first starts off as a single node {\it living-room}, and gradually appends zero, one or more rooms of certain types to the leaf nodes. The probability distribution that we use is shown in Tab.~\ref{tab:room-children} and Tab.~\ref{tab:room-children-upstairs}.
\begin{table}[!h]
    \centering
    \begin{tabular}{ll|l}
        \toprule
         Room parent& Room children & Probability\\
         \midrule
         LivingRoom & LivingRoom & $0.1$\\ 
          & Bedroom & $\mathrm {Cat} (0, 0.3, 0.3, 0.3, 1)$\\
          & Closet & $0.1$\\
          & Bathroom & $0.4$\\
          & Garage & $0.4$\\
          & Balcony & $0.2$\\
          & DiningRoom & $0.8$\\
          & Utility & $0.2$ \\
          & Hallway & $\mathrm{Cat}(0.5, 0.4, 0.1)$\\
            Kitchen & Garage & $0.1$\\
            &Utility & $0.2$\\
            Bedroom & Bathroom & $0.3$\\
            &Closet & $0.5$\\
            Bathroom & Closet & $0.2$\\
            DiningRoom & Kitchen & $1.0$ \\ 
            & Hallway & $0.2$ \\
         \bottomrule
    \end{tabular}
    \caption{Probability of the number of rooms PCFG produces for each leaf node in the graph for the ground floor. Such probability is conditioned on the parent room type (Column 1) and the children room type (Column 2). The probability (Column 3) can either be a Bernoulli distribution (shown as the sole parameter) or a Categorical ($\mathrm {Cat}$) distribution (shown as the probability of the number of children, starting with zero). }
    \label{tab:room-children}
\end{table}
\begin{table}[!h]
    \centering
    \begin{tabular}{ll|l}
        \toprule
         Room parent& Room children & Probability\\
         \midrule
         LivingRoom & Bedroom & $\mathrm {Cat} (0, 0.4, 0.5, 0.2)$\\
          & Closet & $0.2$\\
          & Bathroom & $0.4$\\
          & Balcony & $0.4$\\
          & Utility & $0.2$ \\
          & Hallway & $\mathrm{Cat}(0, 0.5, 0.5)$\\
            Bedroom & Bathroom & $0.3$\\
            &Closet & $0.5$\\
            Bathroom & Closet & $0.2$\\
            Balcony & Utility &0.4\\
            & Hallway & 0.1\\
         \bottomrule
    \end{tabular}
    \caption{Probability of the number of rooms PCFG produces for each leaf node in the graph. The annotations are similar to Tab. \ref{tab:room-children} }
    \label{tab:room-children-upstairs}
\end{table}
Additional edges are added to rooms to create a floor plan graph based on the generated tree. Additional hallways are added and shared among the children with the same parent. Based on the number of floors and the current level, a porch ({\it balcony}) or {\it staircase} may also be added to the graph. All room plans that do not observe bathroom privacy (i.e. a bedroom is connected to a bathroom without going through other bedrooms) or are not planar are rejected. Based on the user input, floor plans with an incorrect number of designated rooms are also rejected. By default, we require all floor plan graphs to have at least one living room and one bathroom. 

\subsection{Floor plan initialization}
Based on the floor plan graph for a specific floor, we first deduct an estimated contour area based on the sum of typical areas of all the rooms on one floor, which can also used to derive the width and length of the contour. To derive the contour on one floor, we randomly bevel the corners with a rectangular, round, or 45-degree profile that provides the diversity of the contour shape. Contours for floors upstairs are either the exact same copy of the contour on its lower floor or a subset of the contour on its lower floor. 

The spaces are subdivided from the contour following the Mondrian Process \cite{Roy2008TheMP}. For each iteration, we randomly select a mostly rectangular space and divide it along one of its axes, and we repeat such division so that there are 1.5 times more blocks than are required in the floor plan graph. All divisions apply by rounding off the division onto a grid with a size of $0.5$, and divisions leading to a bad aspect ratio are rejected. We merge the spaces until there's the same number of spaces as in the floor plan graph, then compute the adjacency relations of all divided spaces, where spaces are adjacent if they share an edge of size greater or equal to a threshold (to place doors). We randomly add a staircase placeholder inside the contour for multistory floor plans, which roughly indicates the location of the staircase. The staircase placeholder ensures staircases across adjacent floors are in the same spatial location. 

Among these contour divisions, we try to find one where the assignment of rooms suffices the floor plan graph via adjacency relations. In addition to the adjacency relations in the floor plan graph, we also ensure that all exterior-facing rooms, including the bedroom, garage, and balcony, have access to the house's exterior. Only the divided spaces intersecting with the staircase placeholder can be assigned to the staircase room. We can find a proper assignment of rooms that satisfies the floor plan graph and other constraints via depth-first search.

\subsection{Objective function for floor plan optimization}
The objective function is defined on a floor plan where spaces are assigned to a node in the floor plan graph. The objective is composed of twelve constraints detailed as follows:
\paragraph{Shortest path to entrance} constraint encourages unidirectional room access from the entrance. We compute the shortest path from all nodes to the floor's entrance, either the front entrance for the ground floor or the staircase for rooms upstairs. The path is computed in an axis-aligned fashion and can only traverse connected rooms on the floor plan graph. The amount of detour for each path is the percentage of the path in the wrong direction of the Euclidean distance from the entrance to that room. The objective function is computed as squared detours summed across rooms. Denote $\mathcal F$ as the set of all floors, $e_f$ is the entrance on floor $f$, and $\rightarrow$ is the path allowed by the adjacency between rooms:
\[\mathcal L_{sp}=\sum_{f\in {\mathcal F}}\sum_{r\in f}\left(\frac{\|e_f\rightarrow r\|_{1,\text{direct}}}{\|e_f\rightarrow r\|_1}-1\right)^2\]

\paragraph{Typical room area} constraint encourages room of typical area so that the spaces serve the best function. A list of the typical area occupied by rooms is listed in Tab.~\ref{tab:typical-area}, which is based on a typical US household. The ideal proportion of a room is computed as the typical area of that room divided by the sum of all the room's typical areas on that floor.  The objective function is computed by the difference between a room's ideal proportion on one floor and the room's true proportion on that floor, summed across rooms. For all rooms $r\in f$, we compute its ideal area as 
\[\overline {area}_r = \frac{\mathrm {typical\_area_r}}{\sum_{r'\in f}\mathrm{typical\_area}_r}area_f\]
A formula for the objective function is 
\[\mathcal L_{ta}=\sum_{f\in {\mathcal F}}\sum_{r\in f}\max\left({\frac{\overline{ \mathrm{area}}_r}{\mathrm{area}_r},\frac{\mathrm{area}_r}{\overline{ \mathrm{area}}_r}}\right)\]

\begin{table}[!h]
    \centering
    \begin{tabular}{l|l}
    \toprule
        Room type & Typical area \\
        \midrule
        Kitchen & 20\\
        Bedroom & 20\\
        LivingRoom & 25\\
        DiningRoom & 20\\
        Closet & 3\\
        Bathroom & 7 \\
        Utility & 3\\
        Garage & 35 \\
        Balcony & 6\\ 
        Hallway & 6\\
        Staircase & 20 \\
        \bottomrule
    \end{tabular}
    \caption{Typical area occupied by rooms}
    \label{tab:typical-area}
\end{table}

\paragraph{Room aspect ratio} constraint encourages rooms of certain types to be square. The objective function is computed as the difference between the true aspect ratio and one, squared and summed across rooms. Denote by $\mathcal R_s$ the rooms needed to be square, we have 

\[\mathcal L_{ar}=\sum_{f\in {\mathcal F}}\sum_{r\in f\cap\mathcal R_s}\left(\max\left(\frac{\mathrm{height}_r}{\mathrm{width}_r},\frac{\mathrm{width}_r}{\mathrm{height}_r}\right)-1\right)^2\]

\paragraph{Room convexity} constraint encourages rooms to be overall convex. The convexity of each room is computed as the ratio between the area of the convex hull of a room and the area of the room itself. The objective function is computed as the squared difference between the convexity of a room and one, summed across rooms. 

\[\mathcal L_{conv}=\sum_{f\in {\mathcal F}}\sum_{r\in f}\left(\frac{\mathrm{area}_{\mathrm{convex\_hull}(r)}}{\mathrm{area}_r}-1\right)^2\]

\paragraph{Room wall conciseness} constraint encourages rooms to have fewer boundary edges, which allows rooms to have better-formed geometry along many iterations of perturbations. The objective function is the squared difference between the number of boundary edges of a room with four (minimal number of edges), summed across rooms. 

\[\mathcal L_{conc}=\sum_{f\in {\mathcal F}}\sum_{r\in f}\left(\|{w\in \mathrm{walls}_r}\|-4\right)^2\]

\paragraph{Functional Room area} constraint incentivizes the available area useful for dwellings of the people inside the house, characterized by functional rooms. Functional rooms include kitchens, bedrooms, living rooms, bathrooms, and dining rooms. The objective function is computed as the proportion of the area covered by these rooms, measured in squared distance with one. Denote by $\mathcal R_f$ the set of functional rooms, we have

\[\mathcal L_{func}=\sum_{f\in {\mathcal F}}\left(\frac{\sum_{r\in f\cap \mathcal R_{f}}\mathrm{area}_r}{\mathrm{area}_f}-1\right)^2\]

\paragraph{Room collinearity} constraint incentivizes walls of multiple rooms to be collinear for aesthetics and construction purposes. The objective function measures the number of distinct X or Y coordinates for all walls of rooms across one floor.

\begin{align*}
    \mathcal L_{col}=\sum_{f\in {\mathcal F}}\left(\left|\{x|\exists r\in f, x\text{ the x-coords of a wall in } r\}\right|\right.
    \\
    \left.\left|\{y|\exists r\in f, y\text{ the y-coords of a wall in } r\}\right|\right)
\end{align*}

\begin{figure}[!h]
    \centering
    \includegraphics[width=\linewidth]{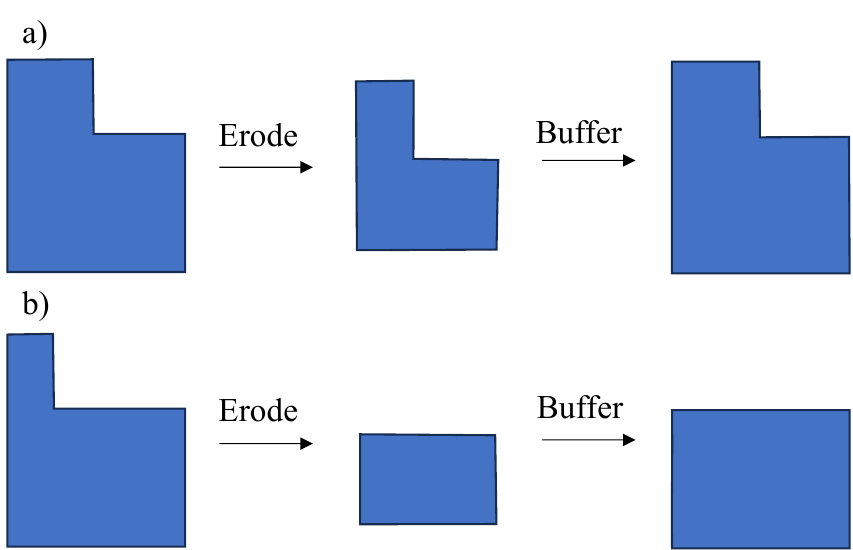}
    \caption{Illustration of narrow passage and erosion-buffer operation. a) In rooms with no narrow passage, the room contour is restored after the operation; b) In rooms with narrow passage, the narrow passage disappears from the room contour after the operation.}
    \label{fig:narrow-passage}
\end{figure}

\paragraph{Narrow passages} constraint limits the number of passages in a room (including hallways) where people or furniture may find it hard to move across. We identify a narrow passage in a room by eroding and then buffering the 2D room contour with a certain threshold margin. Narrow passages inside a room are no longer present in the room contour after that erosion-buffer operation. The objective function is measured as the difference between the area of the room contour pre- and post-erosion-buffer operation. An illustration of the erosion-buffer operation can be found in Fig.~\ref{fig:narrow-passage}. The formula can be written as 

\[\mathcal L_{nar}=\sum_{f\in {\mathcal F}}\sum_{r\in f}\left(\mathrm{area}_r-\mathrm{area}_{\text{erosion-buffer}(r)}\right)\]

\paragraph{Exterior length by room} constraint encourages rooms of certain types to cover most of the exterior walls and windows since people would expect more views of outside in these rooms and more privacy concerns in other rooms. The exterior room types include bedrooms and balconies, denoted by $\mathcal R_e$. The objective function is evaluated using the exterior length covered by these rooms divided by the exterior length covered by all rooms on that floor, measured by its squared distance with one.

\[\mathcal L_{extr}=\sum_{f\in {\mathcal F}}\left(\frac{\sum_{r\in f\cap \mathcal R_e}\sum_{w\in \mathrm{walls}_r\text{ exterior}}\|w\|}{\sum_{r\in f}\sum_{w\in \mathrm{walls}_r\text{ exterior}}\|w\|}-1\right)^2\]

\paragraph{Exterior corner by room} constraint encourages the aforementioned room types to cover most exterior corners, which supposedly have better views. The objective function is measured by the percentage of corners covered by these rooms, measured by its squared distance with one.

\[\mathcal L_{extc}=\sum_{f\in {\mathcal F}}\left(\frac{\sum_{r\in f\cap \mathcal R_e}\left|\{c\in \mathrm{corners}_r|c\text{ exterior}\}\right|}{\sum_{r\in f}\left|\{c\in \mathrm{corners}_r|c\text{ exterior}\}\right|}-1\right)^2\]

\paragraph{Staircase occupancy} constraint encourages the room assigned as the staircase room to cover the staircase placeholder space. It is measured as the percentage of staircase placeholder space covered by the staircase room, measured by its squared distance with one. Denote by $sp$ the staircase placeholder and $\mathcal R_s$ the staircase rooms, we have

\[\mathcal L_{stair\_occ}=\sum_{f\in {\mathcal F}}\left(\sum_{r \in f\cap \mathcal R_s}\frac{\mathrm{area}{sp\cap r }}{\mathrm{area}_{sp}}-1\right)^2\]

\paragraph{Staircase IOU} constraint further encourages the room assigned as the staircase room to be exactly the same size, shape, and location as the staircase placeholder space. It is measured as the IOU of the staircase placeholder with the staircase, measured by its squared distance with one. 

\[\mathcal L_{stair\_occ}=\sum_{f\in {\mathcal F}}\left(\sum_{r \in f\cap \mathcal R_s}\mathrm{IOU}_{r,sp}-1\right)^2\]

\subsection{Floor plan optimization moves}
\begin{figure}[!h]
    \centering
    \includegraphics[width=\linewidth]{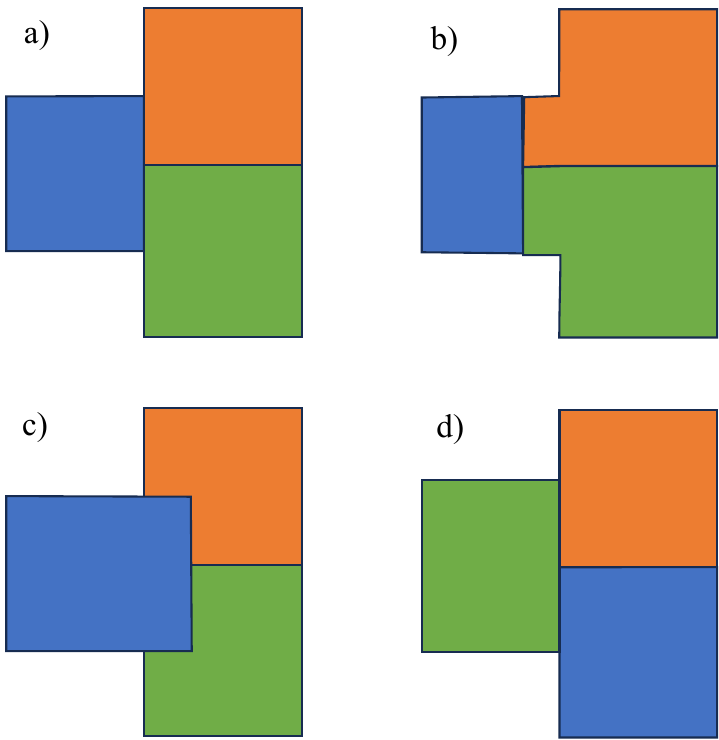}
    \caption{All floor plan optimization moves. a) The original floor plan, with each color showing the assignment of each room (e.g. blue for {\it living-room\_0}, orange for {\it bedroom\_0}, and green for {\it bedroom\_1}; b) floor plan after extruding rightmost wall segment of the blue room inwards; c) floor plan after extruding rightmost wall segment of the blue room outwards; d) floor plan after swapping the assignment for the green and the blue room.}
    \label{fig:moves}
\end{figure}

While solving for the aforementioned constraints, we need to design a set of moves to perturb the floor plan, which are listed as follows and illustrated in Fig.~\ref{fig:moves}:
\paragraph{Extruding a wall segment inwards}
randomly select one wall segment of a room in the current floor plan and move it towards the inside of the room by one grid size (0.5). Other rooms sharing part of the wall with the selected wall will fill up the space left by the move.

\paragraph{Extruding a wall segment outwards}
randomly select one wall segment of a room in the current floor plan and move it towards the outside of the room by one grid size (0.5). Other rooms sharing part of the wall with the selected will give up their space to the room.

\paragraph{Swapping the assignment for adjacent rooms} randomly select one space for a room and its neighbor and swap their room assignment. 

In all of the above moves, we reject moves that lead to a floor plan that does not suffice the floor plan graph on that floor. We also reject moves that lead to invalid geometry, including degenerate, disconnected, or out-of-boundary rooms, and those that fail to satisfy the constraints on exterior rooms and staircase placeholders. One may think of satisfying floor plan graphs as a hard constraint.

\paragraph{Moving staircase.} We also provide an additional move for the staircase placeholder. The staircase placeholder can move along one of the axes by one grid size. 

At each iteration of the simulated annealing, we first select one of the floors to operate on or choose to move the staircase placeholder. Then, we randomly choose one of the three moves to apply. A move is rejected if it no longer satisfies the hard constraint given by the floor plan graph or rejected by the simulated annealing probability computed using the change in the objective function. 

\subsection{Postprocessing of floor plan}
After acquiring the floor plan for all floors, we must convert it to a mesh. Each space assigned to a room is extruded by the height of the wall and solidified by the thickness of the wall; both parameters are the same across all rooms. Besides placing furniture, we conduct the following postprocessing operations.

\paragraph{Placement of doors and windows.} For pairs of rooms that share an edge in the floor plan, they must share a wall segment with its length over a certain threshold. We then cut the shape of the door from both room meshes and place the door in that space. Doors can be opened towards the inside of the house or away from the house's entrance for ergonomics. For other pairs of rooms designated by the user, i.e., between dining rooms and living rooms, one may choose to remove all the walls in between and place no doors. For rooms facing the exterior of the house, if they can have windows installed, we selectively cut off the shape of the window from the room meshes with a limit on the maximal width of the window. Then, we place windows in these shapes. Landscapes are placed outside the window. 

\paragraph{Adding materials to floors, ceilings, and walls.} The walls of rooms are applied with the following materials conditioned on the room type: (ceramic) square tile,  concrete, brick, or plaster. The floors of rooms are applied with the following materials conditioned on the room type: tiled wood floors, square or hexagonal, alternating or non-alternating tiles, rug or concrete. The ceilings of rooms are painted with plaster. Materials are sometimes shared across different rooms.

\paragraph{Adding staircases. } We compute the intersection of the space assigned as staircase rooms on consecutive floors. Our constraint solver will make sure that the intersection is at least the size of the staircase placeholder, which is non-empty. We randomly sample one staircase per floor (excluding the topmost one) and position them inside their corresponding staircase rooms. We reject samples where the staircase and the room in front of the steps fall outside the room or when the consecutive staircase intersects. We cut off the shape of the stairs from the room meshes and add guard rails around the stairs. 

\section{Constraint Solver Details}

\begin{algorithm}
  \caption{Greedy Solving Algorithm}\label{alg:greedy_solver}
  \begin{algorithmic}[1]
    \Procedure{GreedySolver}{$P$}
    \State $\Call{SimulatedAnnealing}{P, Rooms, r}$
    \For{$r$ in $rooms$}
        \State $\Call{SimulatedAnnealing}{P, BigObjects, r}$
        \State $\Call{SimulatedAnnealing}{P, MediumObjects, r}$
        \State $\Call{SimulatedAnnealing}{P, SmallObjects, r}$
    \EndFor
    \EndProcedure
  \end{algorithmic}
\end{algorithm}

\subsection{Greedy Solving Algorithm}
Optimizing over all rooms and all objects at the same time is unfeasible due to the magnitude of the state space. As a result, we use a greedy algorithm to first solve the floor plan, then solve large, medium, and small objects, respectively. At each stage we solve each room separately. A very high-level pseudocode of our solver algorithm is given as Algorithm \ref{alg:greedy_solver}. This algorithm is not optimal in any sense, but the problem at hand is computationally intractable, and an optimal solution is not required to obtain aesthetically pleasing scenes. We provide this solver to prove our language can be optimized efficiently, and to serve as a baseline for future improvements or follow-up work.  

\subsection{Move Utilities}
\label{sec:utilites}

\paragraph{Cardinality Bounding}
Our solver starts with an empty scene, and must add objects during optimization to satisfy object-quantity constraints and objectives given by the user. We implement this via the \textit{Addition} and \textit{Deletion} moves described below, which add or remove one object. Choosing to propose a random object type with a random set of relations would have a vanishingly small likelihood of producing a move that obeys the given constraints - typically only a few object types and a few relation assignments (against wall, on floor, etc) are actually valid. 

To optimize efficiently, we implemented a recursive procedure to traverse through the constraint graph and find every relevant context (such as "on top of bookcase" or "against livingroom wall") available in the current scene state, retrieve any lower/upper bounds on object counts to be placed into these contexts. For example, if there are two shelves in the current state, and the user has specified each shelf shall have between 1 and 5 books placed on it, our procedure would return 2 bounds, one for each shelf, with 1 and 5 as the lower and upper bounds on object count. 

These bounds are sensitive to the scene's current state: if the user specifies there should be more chairs in the dining room than tables in the dining room, then the current number of chairs will be used as an upper bound for the number of tables and vice versa. This allows optimization of arbitrary inequalities between object counts, since by randomly performing valid additions and deletions, the optimizer will explore the full space of discrete object counts for every possible context. 
\begin{figure*}[!h]
    \centering
    \includegraphics[width=\linewidth]{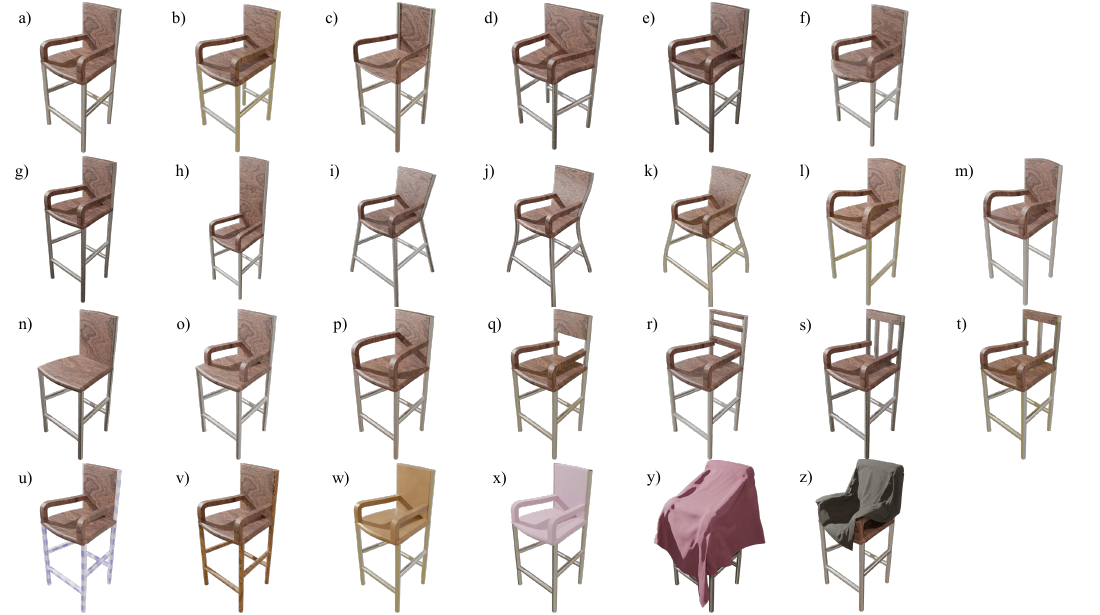}
    \caption{Variation in a chair asset with tuneable parameters. a) A base chair for comparison, followed by chairs with b) larger depth; c) thinner seats; d) wider backs; e) more curvature in seats; f) extrusion in front; g) longer legs; h) larger backs; i)-k) tilted / outward-bending / inward-bending legs; l), m) no leg bars along both axes; n) no arms; o)-p) arms with different attachments to the seat and back; q)-t) backs partially covered/supported by horizontal or vertical bars; u)-v) different leg material (woven fabric/wood); w-x) different seat material (leather/fabric); y)-z) different placement of blankets. }
    \label{fig:variation}
\end{figure*}
\paragraph{Degree Of Freedom Computation}
Moving objects in the full 6D pose space is completely unfeasible because of the plane assignment's hard constraints that need to be satisfied. Enforcing these constraints by minimum distance scores and considering movement only on the XY plane is another option, but this still causes a violation of the hard constraints and is also wasteful as an optimizer state-space. Therefore, we calculate the degrees of freedom of each object and only move objects along the allowed subspace (e.g. painting only moves on the wall it is assigned to).

For each object, we first obtain the planes that the object is constrained to move on. We then compute two types of DOFs. The translation DOFs are computed as the matrix of projection onto the intersection subspace of the planes. If the constraints are contradictory, this will be the zero matrix. The rotation DOF is either the free axis of rotation around which the object is allowed to rotate, or none if the constraints do not allow rotational movement. 

\paragraph{Resolving Discrete Move Poses}
The optimizer needs to initialize every object that is added to the scene before proposing any moves to it. This initialization must obey the plane assignment hard constraints, so that the subsequent continuous moves also obey the hard constraints. Thus, we initialize objects by essentially sampling a random position on the subspace defined by the plane assignments and sampling a random rotation that is a multiple of $\pi/2$. The position sampling is done by sampling a random position on the first plane, and then repeatedly snapping the object to its assigned planes with the specified margin. The validity of the initialization is checked after each attempt, and if each initialization attempt is unsuccessful for a certain number of attempts (20 by default), the initialization is unsuccessful, and the move is reverted.

\paragraph{Reversing Moves}
Not every proposed move is a valid move. For instance, translating a painting too much might cause it to overhang, or reassigning a sofa to another wall might make it intersect with another object in the scene. As a result, after we apply any move, we check its validity. This is done by checking that the chosen object does not collide with any other mesh, and that all the relation constraints of the object are satisfied. If the move is not valid, then we reverse the move to remove its effects. For instance, if the object was moved by a rotation or translation, and the resulting state is not valid, we restore the backup pose of the object. If the move was an addition, we remove the object, and so on.

\begin{figure*}[ht]
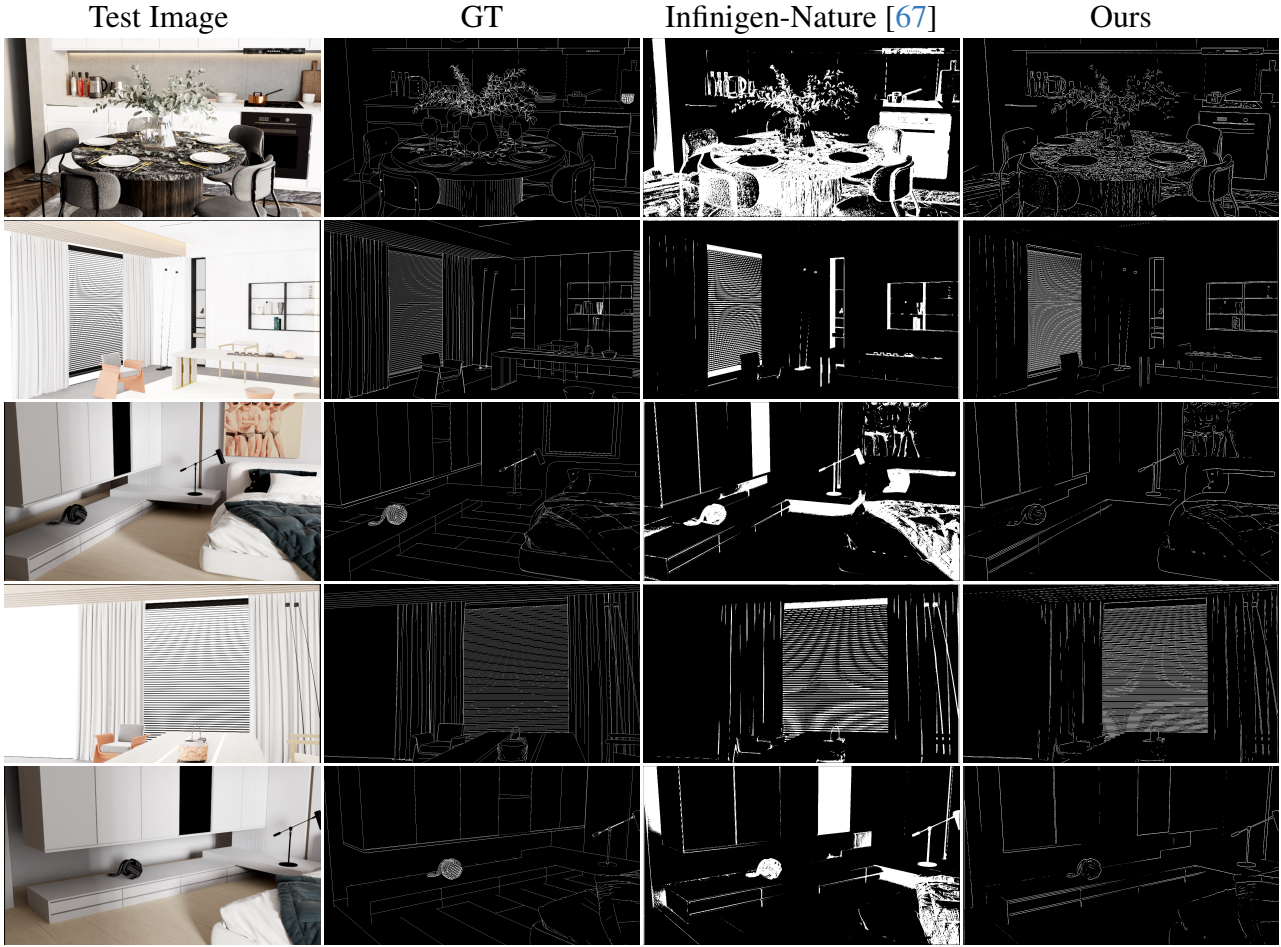

  \begin{center}
  \resizebox{\textwidth}{!}{
    \setlength{\tabcolsep}{0.6pt}
    \begin{tabular}{ccccc}
      Test Image &
      GT &
      Infinigen-Nature~\cite{infinigen2023infinite} &
      Ours & 
 \\
    \galleryRowCompare{figures/occlusion_supp}{media_images_0.jpg}
    {media_images_0_gt.jpg}
    {media_images_0_out.jpg}
    {media_images_0_in.jpg}
    \galleryRowCompare{figures/occlusion_supp}{media_images_1.jpg}
    {media_images_1_gt.jpg}
    {media_images_1_out.jpg}
    {media_images_1_in.jpg}
    \galleryRowCompare{figures/occlusion_supp}{media_images_4.jpg}
    {media_images_4_gt.jpg}
    {media_images_4_out.jpg}
    {media_images_4_in.jpg}
    \galleryRowCompare{figures/occlusion_supp}{media_images_2.jpg}
    {media_images_2_gt.jpg}
    {media_images_2_out.jpg}
    {media_images_2_in.jpg}
    \galleryRowCompare{figures/occlusion_supp}{media_images_3.jpg}
    {media_images_3_gt.jpg}
    {media_images_3_out.jpg}
    {media_images_3_in.jpg}

    \end{tabular}
}
  \end{center}
  \vspace{-1.5em}
  \caption{Additional qualitative results on synthetic scenes~\cite{gumroad}.}
  \label{fig:occlusion_qual_add}
\end{figure*}

\subsection{Move Implementations}

\paragraph{Addition}
To perform an addition, we extract all available cardinality bounds from the current. Then, we discard all bounds that are tight above, i.e., those for which adding an object would violate an upper bound. 

The most challenging stage of addition is finding a satisfying assignment for relevant constraints. If the user specified \texttt{scene[Seating].related\_to(room, on\_floor).related\_to(room, against\_wall).count() > 0}, then we know we must add some kind of seating that is against both the floor and wall. However, many options exist: Seating could mean either an armchair or a sofa, there are many possible floors to place the seating onto, and many possible wall planes attached to each floor plane. Moreover, any choice for these variables could activate additional constraints; for example, the user may have written a rule that applies to all sofas in the scene, or all objects in a particular room, so if we choose for our seating object to be a sofa, or if we choose to put it in that particular room, then additional constraints may be added to the list yet to be satisfied. 

This relation assignment problem is related to classic SAT solving, except for that making an assignment can add additional terms to the equation. Alternatively, it is an SAT problem where the full equation to be satisfied is deceptively long due to new constraints being activated. We anticipate that future versions of our solver can directly incorporate classic SAT-solving approaches. However, for our current constraint programs, we have found it is sufficient to perform

exponential search over all options, visiting each child node in the search tree in a random order to ensure unbiased results. This approach is exponential in the number of semantic and relationship constraints involved, but fortunately, these rarely number more than 3 or 4 (IE, 1-2 semantic classes ), and the branching factor tends to be small (IE, relatively few different specific object options, or few different wall planes to assign to).

For each valid assignment found, we procedurally generate a "placeholder" asset and attempt to fit it into the scene as described in "Resolving Discrete Moves" as described above. Placeholders are special versions of our 3D assets provided by each procedural generator which have mostly planar surfaces and lower polygon count. E.g. the placeholder for a chair would still have properly shaped legs, seat and backrest, but would not have any bevels, chamfers, nails/screws or fine geometric details. This lower polygon representation speeds up collision checking, and eliminates the need for us to heuristically detect flat planes on the object, since the asset author provides these procedurally.

\paragraph{Deletion}
Deletion uses the same cardinality bound logic as addition, but chooses a random object cardinality bound that is not tight below, and proposes to delete it to see if the score is reduced. Typically, these moves do not help the immediate score, as we incentivize placing as many objects as possible, but they can help to eliminate particularly poorly placed objects or to avoid local optima in object counts.

\paragraph{Resample}
Resample's primary function is to replace an existing object in the scene with an object of the same class but with new parameters. This often causes a change in shape, e.g. the length/width of a table will change, or the number of cells in a shelf may increase. Changing these parameters is desirable as it may increase/decrease the objective function (e.g. if a volume() or min\_distance) is changed as a result). To place the new object in the scene, we try aligning each of the bottom corners of the new bounding box with that of the old object, and check each pose for collisions, which allows the object to grow/shrink strictly to the left or right if it is attached to a wall. We assume relation assignments from the old object remain valid, since regenerating an object with new parameters does not change its semantics. 

\paragraph{Translate}
Let $\mathrm{P}$ be the projection matrix computed as the translational degree of freedom for the chosen object. We sample $\vec{x} \in \mathbb{R}^3$ where $x_i \sim \mathcal{N}(0, \sigma^2)$ for $i = 1,2,3$, and the variance $\sigma^2$ is proportional to the temperature. The object is then translated by $\mathrm{P}\vec{x}$. This makes the object take a random step along the subspace on which it is constrained. 

\begin{figure}
    \centering
  \resizebox{\linewidth}{!}{
    \includegraphics[width=\textwidth]{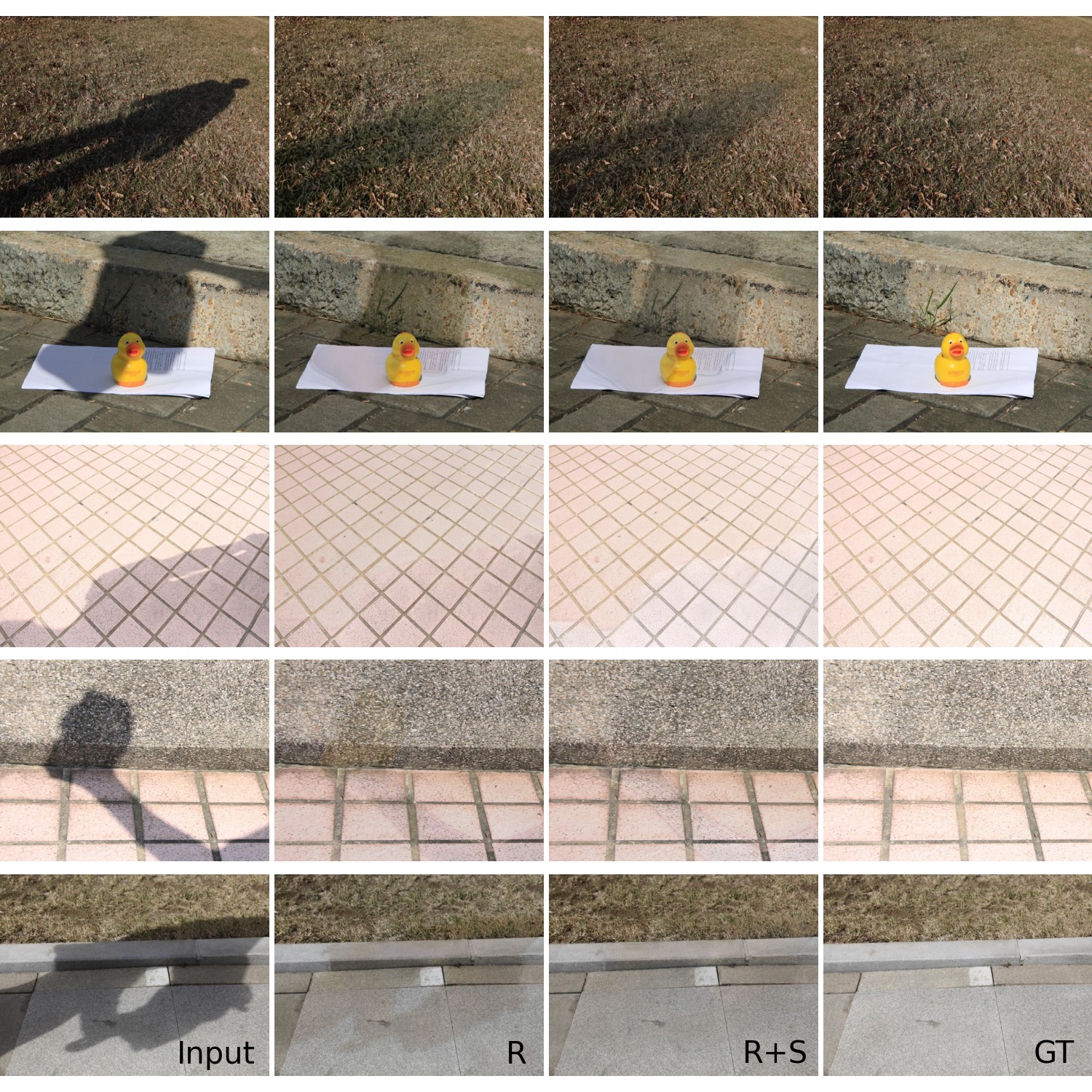}}
    \caption{Additional qualitative zero-shot results on SRD \cite{srd} test dataset.}
    \label{fig:shadow_qual_sup}
\end{figure}

\paragraph{Rotate}
Let $\vec{e}$ be the axis of rotation computed as the rotational degree of freedom for the chosen object. We sample $\theta \sim \mathcal{N}(0, \sigma^2)$ where the variance $\sigma^2$ is proportional to the temperature. The object is then rotated by $\theta$ around the axis $\vec{e}$. This makes the object take a small random rotation on the subspace on which it is constrained. 

\paragraph{ReinitPose} reinitializes the 6DOF pose of the object by resolving the discrete move poses again. Since the object relations are the same, the effect is essentially sampling a random position and orientation on the same constraint subspace. This move is useful for getting a good layout in the early stages of optimization and the cases in which an object is stuck in a sub-optimal position. 

\paragraph{ReassignPlane}. As explained in Addition, if the user specifies an object to be placed against one or more surface(s), then multiple options usually exist for which surfaces to use. This move simply attempts to swap the object to a different plane, e.g. move a sofa to a different wall, or a bottle to a different row of a shelf. 

\paragraph{ReassignTarget} Similarly, multiple options often exist for which object an object is a child of in the scene graph, e.g. a plant pot could rest on one of many different shelves/tables in a room. This move swaps the object to be a child of some other object in the scene that satisfies the same constraints as its current assignment.

\section{Asset Generation Details}
\label{sec:asset_gen}
\subsection{Asset Coverage and Variation}
We provide 79 randomized procedural object generators. By category, we cover Appliances (10 generators, 112 params), Windows/Doors/Staircases (14 generators, 127 params), Furniture (17 generators, 216 params), Decorations (15 generators, 92 params), and Small Objects (19 generators, 194 params). We provide 30 material generators, 120 params total, split approximately evenly between types of wood, ceramic, fabric, metal and others.  Materials are assigned to objects via customizable weighted lists, e.g. spatulas invoke either wood, plastic or metal generators for each of their ends. Following Infinigen, we report procedural parameter count as a proxy of complexity; each parameter is a random but controllable degree of freedom e.g ``number of seats on a sofa". In all, we provide 40k lines of code, of which 25k are object/material generators. For asset coverage, an incomplete list of the assets we cover is listed in Tab.~\ref{tab:coverage}. For asset variations, an illustration of the variation of assets is shown in Fig.~\ref{fig:variation}.

\subsection{Lighting and Camera Placement} Lights and windows use similar constraints as other objects (e.g.\ maximize count \& spacing), with random wattage/temperature sampled from real-world ranges. Camera selection follows Infinigen \cite{infinigen2023infinite}: we sample at random, reject near walls, and maximize depth variance.

\section{Experiment Details}
\label{sec:app_exp}

\subsection{Shadow Removal}
We use the model implementation from \cite{shadowFormer}'s codebase to train the two variants of the model: R (trained on real dataset only) and R+S (trained on real and synthetic datasets). Since the codebase lacked a validation set, we developed our own, comprising all image pairs across four scenes from the ISTD training dataset. Additionally, in contrast to the provided implementation, we used an L1 loss as stated in the paper. We trained the two variants for 30k steps each, including a 30-epoch linear warmup phase, using AdamW \cite{adamw} optimizer with default hyperparameters and a learning rate of $2e-4$. We chose the runs to have an effective batch size of 32 (by accumulating gradients for 4 steps and using the actual batch size to be 8). The training process utilized four Nvidia 3090 GPUs, with Mixed-16 precision. Fig. \ref{fig:shadow_qual_sup} shows additional qualitative results.

We opted not to use the pre-trained model from the codebase, as our attempts to reproduce the results were unsuccessful. Nevertheless, to ensure a fair comparison, we adhered to the same implementation details for both variants. Additionally, we chose not to report SSIM (Structural Similarity Index Measure), since both models demonstrated equivalent performance, with no significant difference observed when rounding to two decimal places, for this metric.

\subsection{Occlusion Boundaries}

We separately train three U-Net~\cite{ronneberger2015u} models from scratch on images generated from \projectname{}, Infinigen~\cite{infinigen2023infinite} and Hypersim~\cite{Roberts_2021_ICCV}. We apply random \{cropping, brightness contrast\} and color jittering with probability $0.6$. We also use the RMSprop optimizer with a base learning rate of $10^{-5}$, a momentum of 0.99 and a weight decay of $10^{-8}$. Each model is trained for 10 epochs using binary cross-entropy loss. 

Due to the absence of ground truth occlusion boundaries in Hypersim (or any other photorealistic dataset), we approximate them by thresholding the gradient of the provided depth maps. We carefully tuned this threshold on Hypersim to give the best results.

We compare the performance of the U-Net models on a curated test set of photo-realistic artist-designed synthetic 3D scenes for architecture visualization~\cite{gumroad}. We extract the ground truth occlusion boundaries of these scenes using the tools provided in Infinigen. Additional qualitative results shown in Fig. \ref{fig:occlusion_qual_add} underscore our claim that the \projectname{} - trained model generalizes better.

\begin{table}[h]
    \centering
    \resizebox{\columnwidth}{!}{
    \begin{tabular}{@{}l|c|c|c|c|c@{}}
    Method &
    \begin{tabular}{@{}c@{}}
    Mean Error \\
    Frequency $\downarrow$
    \end{tabular} &%
    \begin{tabular}{@{}c@{}}
    More $\uparrow$ \\
    Realistic
    \end{tabular} &%
    \begin{tabular}{@{}c@{}}
    More Realistic \\
    Layout $\uparrow$
    \end{tabular} &%
    \begin{tabular}{@{}c@{}}
    Realism \\
    CI $99\%$
    \end{tabular} &%
    \begin{tabular}{@{}c@{}}
    Layout Realism \\
    CI $99\%$
    \end{tabular} \\
    
    \hline
    
    ProcTHOR \cite{Deitke2022ProcTHORLE} & 0.252 & 0.107 & 0.056 & $[0.054,0.187]$ & $[0.021,0.127]$ \\ 
    ATISS \cite{atiss} & 0.232 \cite{atiss} & 0.287 & 0.307 & $[0.198,0.389]$ & $[0.217,0.410]$ \\
    SceneFormer \cite{scene_former} & 0.713 \cite{atiss} & 0.333 & 0.440 & $[0.241,0.439]$ & $[0.339,0.547]$ \\
    FastSynth \cite{ritchie_fast_2018} & 0.414 \cite{atiss} & 0.093 & 0.147 & $[0.046,0.171]$ & $[0.083,0.234]$ \\
    Ours & \textbf{0.175} & \textbf{0.795} & \textbf{0.760} & $[0.750, 0.835]$ & $[0.712, 0.803]$ \\
    \end{tabular}}
    \captionof{table}{\textbf{Perceptual Study Results}. We followed the method and metrics from ATISS, but added \textit{Layout Realism}, which says to only consider arrangement. We used each method's default renderer.}
    \label{tab:ustudy_res}
    \vspace{-0.5em}
\end{table}

\section{Perceptual Study}
\label{sec:perceptual}
Following ATISS \cite{atiss}, we conducted a perceptual study on Amazon Mechanical Turk to evaluate the realism of the generated scenes and the realism of the generated layouts. We compared Infinigen Indoors to ProcTHOR \cite{Deitke2022ProcTHORLE}, ATISS \cite{atiss}, SceneFormer \cite{scene_former}, and FastSynth \cite{ritchie_fast_2018}. We presented the subjects pairs of images from each method (for instance Infinigen vs ProcThor) to evaluate overall realism and layout realism. For mean error frequency, we asked the subjects if the image from a method contained any obvious errors such as flying furniture, overlapping furniture, etc. For layout realism, we asked the subjects to focus only on the arrangement of the furniture and ignore the style of individual objects. Table \ref{tab:ustudy_res} shows that the subjects preferred Infinigen Indoors over all methods in terms of both realism, layout realism, and the lack of obvious errors. An important caveat is that ``realism" may be influenced by asset and lighting quality.

\begin{table}[H]
\begin{itemize}
    \item Household appliances
    \begin{itemize}
        \item Fridge, Beverage fridge (with racks)
        \item Dishwasher (with racks)
        \item Microwave
        \item Oven, Stove, Oven with stove (with racks)
        \item TV, Monitor
        \item Kitchen Sink (with faucets)
    \end{itemize}
    \item Bathroom fixtures
    \begin{itemize}
        \item Bathroom sink (standing / embedded / tabletop)
        \item Bathtub (alcove / freestanding / corner)
        \item Hardware (towel bar / towel ring / toilet roll paper holder / robe hooks)
        \item Toilet (two-piece / one-piece / in-wall)
    \end{itemize}
    \item Clothes
    \begin{itemize}
        \item Pants (underwear / shorts / pants)
        \item Shirts (T-shirts / shirts)
        \item Blankets / towel (folded / rolled)
    \end{itemize}
    \item Architectural Elements
    \begin{itemize}
        \item Doors
        \begin{itemize}
            \item Lite / Louver / panel / glass panel door
            \item Door casings
        \end{itemize}
        \item Staircases (with treads / banisters / guardrails / glass railings)
        \begin{itemize}
            \item Straight / Cantilever / L-shaped / U-shaped staircase
            \item Spiral / curved staircase
        \end{itemize}
        \item Rugs
        \item Warehouse racks / pallets
    \end{itemize}
    \item Seatings
    \begin{itemize}
        \item Bar stool / office chair
        \item Armchair / dining chair / side chair / spholstered chair
        \item Beds (bedframe / mattress / pillow)
        \item Sofa
    \end{itemize}
    \item Shelves (with drawers and doors)
    \begin{itemize}
        \item Cabinets / kitchen cabinets
        \item Cell shelves / wall shelves / bookcases / triangle shelves
    \end{itemize}
    \item Table decorations
    \begin{itemize}
        \item Books (column / stack)
        \item Vases / Aquarium tank
        \item Plants in pots (floor-top / table-top)
    \end{itemize}
    \item Tables
    \begin{itemize}
        \item Desks / Cocktail table / Dining table / Kitchen table
    \end{itemize}
    \item Tableware
    \begin{itemize}
        \item Bottle (soda / wine / beer / juice) / jar
        \item Chopsticks / Knife(table knife / cleaver / chef's knife) / forks / spoons / spatulas / bowl / plate
        \item Cup (mug / shot glass / teacup / plastic cup) / wineglass
        \item Food bag(chip bag / food pouch / food bar) / Food box / can / jar / Fruits in containers(i.e. tableware with fruits placed inside)
        \item Pan / pot(Cooking pot / saucepan) / lid(of pots and pans)
    \end{itemize}
    \item Wall decorations
    \begin{itemize}
        \item Balloons / wall arts / mirror
    \end{itemize}
    \item Windows
    \begin{itemize}
        \item Sliding / awning / casement / glassblock / bay window
    \end{itemize}
\end{itemize}
\caption{Coverage of the assets in \projectname{}.}
\label{tab:coverage}
\end{table}

\end{document}